\documentclass[sigconf,nonacm]{acmart}

\usepackage{booktabs}
\usepackage{multirow}
\usepackage{makecell}
\usepackage{wrapfig}
\usepackage{graphicx}
\usepackage{amsmath}
\usepackage{microtype}
\usepackage{balance}
\usepackage{tabularx}
\usepackage[most]{tcolorbox}
\usepackage{algorithm}
\usepackage{algpseudocode}
\usepackage{capt-of}
\usepackage{placeins}
\usepackage{float}
\usepackage{listings}
\lstdefinestyle{promptstyle}{
  basicstyle=\ttfamily\scriptsize,
  breaklines=true,
  breakatwhitespace=false,
  breakindent=0pt,
  columns=fullflexible,
  keepspaces=true,
  frame=single,
  linewidth=\columnwidth,
  xleftmargin=0pt,
  xrightmargin=0pt,
  framexleftmargin=2pt,
  framexrightmargin=2pt,
  framesep=2pt,
  aboveskip=4pt,
  belowskip=5pt,
  showstringspaces=false
}

\newcommand{\best}[1]{\textbf{#1}}
\newcommand{\second}[1]{\underline{#1}}

\begin{document}

\title[CHARM]{CHARM: A Multimodal Graph Foundation Model with Hierarchical Context Modeling for Zero-Shot Transfer}
\author{Ankang Yang}
\affiliation{%
  \institution{School of Computer Science and Technology, College of Intelligence and Computing, Tianjin University}
  \city{Tianjin}
  \country{China}
}

\author{Jitao Zhao}
\affiliation{%
  \institution{School of Computer Science and Technology, College of Intelligence and Computing, Tianjin University}
  \city{Tianjin}
  \country{China}
}

\author{Di Jin}
\affiliation{%
  \institution{School of Computer Science and Technology, College of Intelligence and Computing, Tianjin University}
  \city{Tianjin}
  \country{China}
}

\author{Yuxiao Huang}
\email{yuxiaohuang@gwu.edu}
\affiliation{%
  \institution{Data Science Program, Columbian College of Arts and Sciences, The George Washington University}
  \city{Washington}
  \state{DC}
  \country{USA}
}

\author{Dongxiao He}
\affiliation{%
  \institution{School of Computer Science and Technology, College of Intelligence and Computing, Tianjin University}
  \city{Tianjin}
  \country{China}
}

\renewcommand{\shortauthors}{Yang et al.}
\hypersetup{
  pdfauthor={Ankang Yang, Jitao Zhao, Di Jin, Yuxiao Huang, and Dongxiao He},
  pdfsubject={Multimodal graph foundation models for zero-shot transfer},
  pdfkeywords={multimodal graphs, graph foundation models, zero-shot transfer}
}

\begin{abstract}

Graph foundation models (GFMs) have emerged as a promising paradigm for transferring knowledge across graph domains and tasks. Real-world graphs associate nodes with text, images, and other modalities, making multimodal graphs essential for representing complex entities and relations. Moreover, collecting labels and adapting models for every new graph domain is costly and often infeasible, motivating zero-shot transfer. Unfortunately, zero-shot transfer on multimodal graphs remains underexplored. Existing GNN-based graph foundation models typically require downstream adaptation, whereas LLM-based graph methods mainly address unimodal graphs or tasks within a single domain. This setting presents two key challenges. First, models must generalize knowledge from individual modalities while capturing transferable cross-modal relations. Second, without target-domain fine-tuning, node representations remain entangled with domain-specific structures and modality-specific characteristics, obscuring shared concepts in unseen domains. To address these challenges, we propose CHARM, a multimodal graph foundation model with hierarchical context modeling for zero-shot transfer. CHARM replaces isolated raw nodes with hierarchical graph contexts that capture multimodal semantics and cross-modal relations. These contexts map domain-specific node patterns to shared high-level concepts, reducing reliance on target-domain supervision or adaptation. A modality-aware graph context encoder integrates multimodal information with graph structure and converts the resulting representations into graph tokens for a large language model (LLM). Experiments show consistent improvements on zero-shot multimodal graph tasks.

\end{abstract}

\maketitle

\section{Introduction}
Graphs provide a unified representation for relational data by modeling entities as nodes and their interactions as edges. They are widely used in social networks, molecular systems, and e-commerce, where neighborhood relations and higher-order dependencies are essential for understanding complex objects and their interactions~\cite{Social_nets,Message-passing,DataAmazon}. Graph neural networks (GNNs) have become an important paradigm for graph learning by propagating information over graph structures to learn task-specific representations~\cite{Survey-GNNS,GCN,GAT,GraphSAGE,GIN}. However, conventional GNNs are typically trained on a particular graph domain and task, causing their learned representations and decision boundaries to be closely tied to specific data distributions and label spaces. This dependence limits their ability to transfer to new domains and tasks. To improve graph transferability, recent studies have explored graph foundation models that aim to acquire reusable structural and semantic knowledge from diverse graphs~\cite{PFMs,GFMSurvey_Position,AnyGraph,OpenGraph}. Existing methods generally follow two directions. GNN-based approaches learn transferable graph representations through pre-training, prompting, or downstream adaptation~\cite{SAMGPT,RiemannGFM,OFA,GCOPE,GFT,MDGPT}, whereas large language model (LLM)-based approaches express graph information as textual descriptions or continuous graph tokens, providing a flexible interface for open-vocabulary prediction and zero-shot task formulation~\cite{GraphGPT,LLaGA,GOFA,HiGPT,Transformer}.

Despite these advances, existing GFMs often focus primarily on single-modality semantics, overlooking the inherently multimodal nature of real-world graphs, where text, images, videos, and other modalities provide diverse yet complementary semantic information~\cite{ektefaie2023multimodal,DBLP:conf/kdd/0004L0YHLZ0W25}. Effectively integrating these modalities is therefore essential for developing a comprehensive understanding of graph entities and their interactions. This has motivated growing interest in multimodal graph learning. Early multimodal GNNs process different modalities through separate propagation or fusion mechanisms, enabling graph structure to interact with modality-specific node information~\cite{DBLP:journals/ijon/WangHQFX20,DBLP:journals/ipm/TaoWWHHC20,MMGPL}. GNN-based multimodal graph foundation models further learn shared multimodal representations across graphs by jointly modeling modality-specific semantics, graph structures, and cross-modal interactions during pre-training~\cite{UniGraph2}. More recently, LLM-based methods integrate textual, visual, and structural information into large-model interfaces, allowing multimodal graph evidence to participate in language-conditioned prediction and reasoning~\cite{DBLP:conf/icml/NingFWXH25,DBLP:journals/corr/abs-2506-02568,DBLP:journals/corr/abs-2603-05181}.

However, zero-shot transfer on multimodal graphs remains largely underexplored and poses substantial challenges. Unlike zero-shot learning on text-attributed graphs~\cite{ZeroG,GraphCLIP,TEA-GLM}, it requires a model to jointly interpret multiple semantic signals from text and images, capture their interactions with graph topology, and generalize across unseen graphs without labeled examples. GNN-based multimodal graph foundation models follow a traditional representation-learning paradigm: although they pretrain unified node representations, downstream prediction still requires labeled data to train a task-specific classifier or adapter. LLM-based graph learning methods reformulate graph prediction as a language-conditioned reasoning problem~\cite{LLM-GNN,LLM-BP,GraphAgent}. However, their zero-shot capabilities have been primarily demonstrated on single-modality graphs, with limited modeling of the interactions among textual, visual, and structural information. The few methods that integrate all three information sources require domain-specific training or graph instruction tuning, leaving the pretrained multimodal reasoning capabilities of large models underexplored for zero-shot graph transfer.

\begin{figure}[t]
    \centering
    \includegraphics[width=\columnwidth]{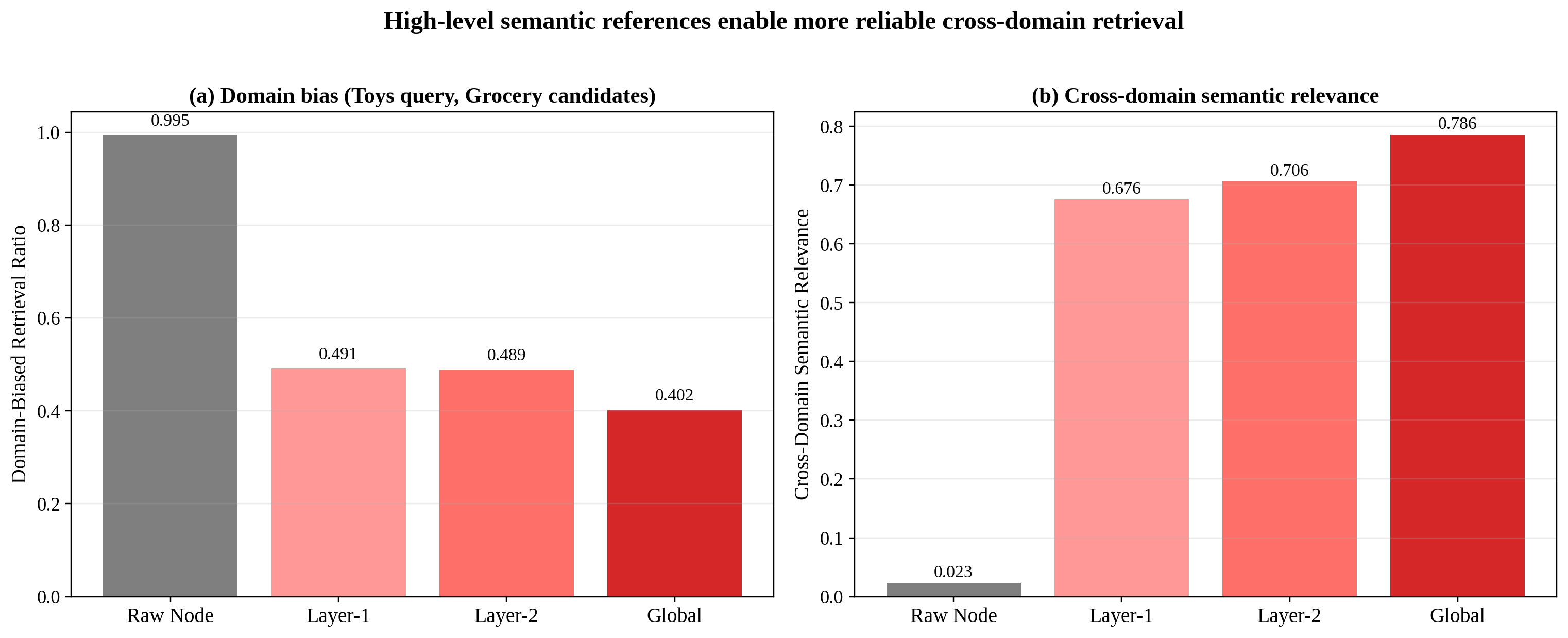}
    \caption{Domain bias and semantic relevance in Toys-to-Grocery retrieval.}
    \label{fig:motivation_retrieval}
    \Description{Two bar charts compare raw nodes, Layer-1 anchors, Layer-2 anchors, and global anchors. Domain bias decreases at higher semantic levels, while Toys-to-Grocery semantic relevance increases.}
\end{figure}

This setting presents two fundamental challenges. First, multimodal graph modeling requires the model to jointly capture graph structure, modality-specific semantics, and interactions across modalities. Different modalities may provide complementary evidence, but their relevance and reliability can vary across nodes and domains. A transferable model must therefore preserve useful knowledge from each modality while identifying cross-modal relations that remain valid under distribution shifts. Second, zero-shot transfer must be achieved without target-domain labels or parameter adaptation. The model can use only unlabeled target attributes and topology to construct prediction contexts. However, raw node features and local neighborhoods are often dominated by domain-specific vocabulary, visual appearance, and connection patterns, making semantically related nodes across graphs difficult to identify. Figure~\ref{fig:motivation_retrieval} further illustrates this challenge through a label-free cross-domain retrieval study in which Toys is the source domain and Grocery is the target domain. Raw-node retrieval yields a domain-bias ratio of 0.995, indicating that nearly all retrieved nodes come from the source domain; among the retrieved target-domain candidates, semantic relevance is only 0.023. These results suggest that instance-level similarity is dominated by domain- and modality-specific details, making shared concepts difficult to identify across unseen multimodal graph domains. 

To address these challenges, we propose CHARM, a multimodal graph foundation model for zero-shot transfer with hierarchical semantic modeling and modality-aware context encoding. Specifically, CHARM first constructs a synthetic hierarchical graph by progressively organizing raw multimodal nodes into Layer-1, Layer-2, and global anchors. These higher-level units provide within-domain and cross-domain semantic references, allowing nodes from different graphs to be associated through shared concepts rather than direct instance-level similarity. As illustrated in Figure~\ref{fig:motivation_retrieval}, moving from raw nodes to higher-level semantic units substantially reduces retrieval domain bias while improving cross-domain semantic relevance, empirically supporting hierarchical abstraction as a more transferable representation unit. CHARM further constructs modality-complementary relations between structurally related nodes whose similarity is evident in either text or vision. For each prediction target, it retrieves a compact graph context containing the center node, structural neighbors, hierarchical semantic references, and modality-complementary neighbors. A graph context encoder then performs reliability-aware multimodal fusion and structural propagation over the retrieved context. The resulting continuous graph tokens encode multimodal semantics, cross-modal relations, and graph structure, and are directly consumed by a frozen LLM together with task instructions.

\begin{itemize}

\item We study zero-shot transfer on multimodal graphs with unseen target domains and label spaces.

\item We construct a synthetic hierarchical graph with Layer-1, Layer-2, and global anchors, together with modality-complementary relations, providing high-level semantic references and cross-modal evidence for zero-shot transfer.

\item We introduce a compact graph context construction strategy and a graph context encoder that convert the center node, structural neighbors, semantic anchors, and bridge neighbors into continuous graph tokens for an LLM.

\item We conduct experiments on zero-shot multimodal graph node classification and link prediction, demonstrating the effectiveness of \textsc{CHARM}.
\end{itemize}

\section{Related Work}
\subsection{Graph Foundation Models}
Recent graph foundation models have explored how to transfer graph knowledge beyond a single dataset or task~\cite{PFMs,GFMSurvey_Position,AnyGraph,OpenGraph}. Early efforts such as SAMGPT and RiemannGFM~\cite{SAMGPT,RiemannGFM} focus on learning reusable structural regularities, using structure tokens, structural vocabularies, or Riemannian geometry to capture common structural patterns across graphs. Other methods such as OFA and UniGraph~\cite{OFA,UniGraph} introduce natural-language descriptions to unify graph tasks and datasets, showing that language can provide a shared semantic interface for graph transfer. More recently, LLM-based graph models such as GraphGPT, LLaGA, and GOFA~\cite{GraphGPT,LLaGA,GOFA} connect graph structures with large language models through graph-text alignment, structure-aware node sequences, or interleaved GNN–LLM architectures, enabling more flexible open-vocabulary prediction and zero-shot task formulation. However, these methods mainly focus on bag-of-words feature graphs or textual attributed graphs, where graph information is represented primarily through structure or language. They pay limited attention to multimodal graphs, where real-world nodes are associated with text, images, and other modalities simultaneously.

\subsection{Multimodal Graph Learning}
Multimodal graph learning studies how to jointly model graph structures and multiple modalities such as text and images. Early methods such as MGCN and MGAT~\cite{DBLP:journals/ijon/WangHQFX20,DBLP:journals/ipm/TaoWWHHC20} extend graph neural networks with modality-specific propagation, fusion, or attention mechanisms, demonstrating the benefit of combining structural dependencies with multimodal attributes. Recently, graph foundation models for multimodal graphs, such as UniGraph2~\cite{UniGraph2}, have improved transferability by aligning multimodal features and graph structures across datasets. However, these methods still largely follow a representation learning paradigm, where downstream prediction depends on task-specific heads or adaptation. With the rise of large models, Graph4MM, MLaGA, and Mario~\cite{DBLP:conf/icml/NingFWXH25,DBLP:journals/corr/abs-2506-02568,DBLP:journals/corr/abs-2603-05181} introduce LLMs or multimodal large models into multimodal graph learning by aligning graph-aware multimodal features with language models, incorporating multi-hop structures, or designing modality-adaptive graph instruction tuning. These methods advance the adaptation of large models to multimodal graph inputs, but they do not fully exploit their zero-shot generalization capabilities across unseen graph domains and differing label spaces.

\begin{figure*}[t]
    \centering
    \includegraphics[width=\textwidth]{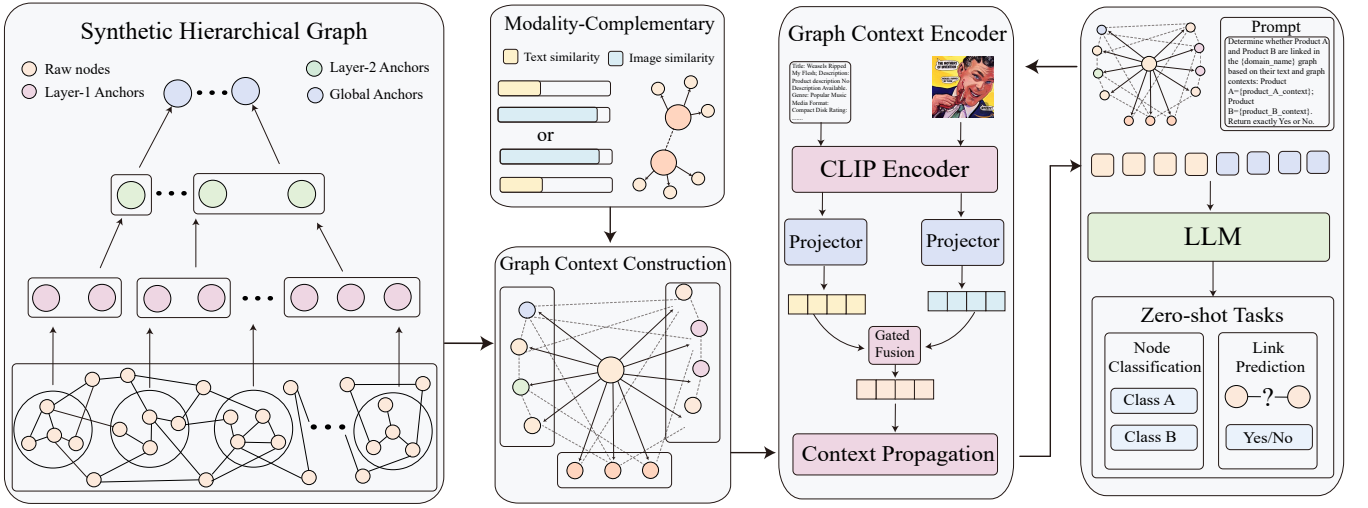}
    \caption{Overview of \textsc{CHARM}.}
    \Description{Overview of the CHARM framework, including hierarchical graph construction, modality-complementary bridge construction, graph context retrieval, multimodal context encoding, and zero-shot prediction.}
    \label{fig:overview}
\end{figure*}
\section{Methodology}
CHARM performs zero-shot prediction on multimodal graphs without target-domain labels or target-domain parameter updates. Its central design is to replace isolated node inputs with hierarchical graph contexts that organize multimodal and structural evidence at different semantic levels. A synthetic hierarchy lifts fine-grained nodes into Layer-1, Layer-2, and global anchors, providing abstract cross-domain references. To model complementary associations that manifest differently across modalities, we introduce modality-complementary relations. For each prediction target, CHARM retrieves a compact context from the original graph, the synthetic hierarchy, and the bridge relations. A graph context encoder then performs reliability-aware multimodal fusion and structural propagation over the retrieved context, producing continuous graph tokens that enable an LLM to make zero-shot predictions.

\subsection{Synthetic Hierarchical Graph}

Raw multimodal nodes describe fine-grained instances whose textual expressions, visual appearances, and local structural patterns can vary substantially across domains. Consequently, direct cross-domain matching is often dominated by domain-specific details rather than the high-level concepts shared across graphs. However, mapping all nodes directly to a small set of global representations would over-compress their semantics and obscure distinctions among individual instances. To balance cross-domain abstraction with instance-level specificity, CHARM constructs a synthetic hierarchy $\mathcal{G}_{\mathrm{syn}}$ that progressively lifts raw nodes into broader semantic units while preserving a path to each original node.

We construct the hierarchy using features extracted by frozen vision and language encoders. For each node $v$, the encoders produce textual and visual features $\mathbf{x}^{t}_v$ and $\mathbf{x}^{i}_v$. We use textual features for hierarchical assignment because product titles and descriptions generally express category-level semantics more explicitly, whereas visual features are often influenced by instance-specific factors such as color, background, and presentation style. Jointly clustering text and image features may make the assignments overly sensitive to domain-specific visual appearance. As a result, the resulting clusters may fail to capture semantic concepts shared across domains. Nevertheless, visual information is not discarded: it is retained and aggregated according to the textual semantics when constructing each synthetic node, ensuring that the semantic anchors at all levels remain multimodal. For each domain $\mathcal{D}_d$, K-means~\cite{mcqueen1967some} partitions the raw nodes into
$K_d^{(1)}$ Layer-1 anchors:
\begin{equation}
\min_{\{\boldsymbol{\mu}_{d,k}^{(1)}\}_{k=1}^{K_d^{(1)}},\,c}
\sum_{v\in\mathcal{V}_d}
\left\|
\mathbf{x}^{t}_{v}
-
\boldsymbol{\mu}_{d,c(v)}^{(1)}
\right\|_2^2,
\end{equation}
where $\mathcal{V}_d$ denotes the raw-node set in domain
$\mathcal{D}_d$, $\mathbf{x}^{t}_{v}$ is the textual feature of node
$v$, $\boldsymbol{\mu}_{d,k}^{(1)}$ is the center of the $k$-th
Layer-1 cluster, and
$c:\mathcal{V}_d\rightarrow\{1,\ldots,K_d^{(1)}\}$ denotes the cluster
assignment.

The resulting Layer-1 anchors capture relatively fine-grained semantic groups within each domain. We recursively cluster their textual representations to obtain Layer-2 anchors, which summarize broader in-domain semantic regions. The Layer-2 anchors from all domains in the same graph group are then jointly clustered into global anchors, yielding the semantic chain
\begin{equation}
v\rightarrow a^{(1)}\rightarrow a^{(2)}\rightarrow p.
\end{equation}
Layer-1 anchors remain close to individual instances, Layer-2 anchors
capture broader domain-level concepts, and global anchors provide a shared semantic reference space across domains. This multi-level
organization avoids directly compressing raw nodes into
a small global vocabulary, which may merge unrelated instances, while
also overcoming the lack of cross-domain references in purely
local clustering.

To make these semantic units available for context propagation, we connect adjacent levels with bidirectional edges. Each raw node is linked to its assigned Layer-1 anchor, and each Layer-1 anchor is linked to its assigned Layer-2 anchor. Each Layer-2 anchor is further connected to its top-$m$ nearest global anchors according to cosine similarity, including its assigned global anchor. These links allow semantically related regions from different domains to interact at the global-anchor level even when direct raw-node matching is unreliable. During zero-shot inference, a target node can therefore access shared high-level semantics through its hierarchical path while retaining its original instance-level information.

For each synthetic node, we construct its textual and visual features separately. Directly averaging member-node text embeddings may blur semantic distinctions and provide little interpretable content for the LLM. We therefore construct a deterministic, label-free semantic summary $s_C$ using TF-IDF~\cite{sparck1972statistical} and fixed templates. Layer-1 summaries use the top eight 1--3-gram terms extracted from at most 80 member titles and descriptions, with the first two terms forming the semantic label. Layer-2 summaries apply TF-IDF to their child summaries and combine frequent child labels, while global anchors use the same rule with a cross-dataset template. No LLM or category label is used. The summary is then encoded by the same frozen text encoder used for raw nodes:
\begin{equation}
\mathbf{a}^{t}_{C}=\mathrm{Enc}_{t}(s_C).
\end{equation}
We then use this semantic summary as a query for text-guided visual aggregation:
\begin{equation}
\mathbf{a}^{i}_{C}=\sum_{v\in C}\alpha_v\mathbf{x}^{i}_{v},\qquad
\alpha_v=\frac{\exp(\cos(\mathbf{a}^{t}_{C},\mathbf{x}^{t}_{v})/\tau)}
{\sum_{u\in C}\exp(\cos(\mathbf{a}^{t}_{C},\mathbf{x}^{t}_{u})/\tau)},
\end{equation}
where $\mathbf{a}^{i}_{C}$ denotes the visual feature of synthetic node $C$, $\mathbf{x}^{i}_{v}$ is the image feature of member node $v$, and $\alpha_v$ is its aggregation weight. The weight is computed from the cosine similarity between the textual feature of the synthetic node $\mathbf{a}^{t}_{C}$ and the textual feature of member node $\mathbf{x}^{t}_{v}$. Members whose descriptions agree more strongly with the cluster meaning receive larger visual weights. In this way, the textual summary defines the semantic identity of the synthetic node, while the visual aggregation supplies representative appearance evidence. The same construction is applied recursively, so Layer-1 anchors, Layer-2 anchors, and global anchors all contain aligned text and image features and can be processed by the same context encoder as raw nodes. Hierarchy formation uses only attributes and assignments.

\subsection{Modality-Complementary Bridges}

The synthetic hierarchy provides semantic abstractions at multiple
resolutions, but it may overlook raw-node associations that are expressed strongly in only one modality. For example, two products may share a similar textual purpose despite having different appearances, or exhibit similar visual forms despite being described with different vocabulary. Requiring text and image similarities to be simultaneously high would discard such one-modality evidence. CHARM therefore constructs modality-complementary bridges between structurally related nodes that are strongly similar in one modality but weakly similar in the other.

For each candidate pair $(v_i,v_j)$, we compute its textual and visual
similarities as
\begin{equation}
s_t(v_i,v_j)
=
\cos(\mathbf{x}^{t}_{v_i},\mathbf{x}^{t}_{v_j}),
\qquad
s_i(v_i,v_j)
=
\cos(\mathbf{x}^{i}_{v_i},\mathbf{x}^{i}_{v_j}).
\end{equation}
The pair is considered modality-complementary when
\begin{equation}
\mathcal{E}_{\mathrm{br}}
=
\left\{
(v_i,v_j)\ \middle|\
\begin{array}{l}
s_t(v_i,v_j)\ge \tau_t^{h},\ s_i(v_i,v_j)\le \tau_i^{l},\\
\text{or }\
s_i(v_i,v_j)\ge \tau_i^{h},\ s_t(v_i,v_j)\le \tau_t^{l}
\end{array}
\right\}.
\end{equation}
The first condition identifies text-strong/image-weak relations, whereas the second identifies image-strong/text-weak relations. The high and low thresholds are set to the 75th and 35th percentiles, respectively, estimated independently from 200,000 sampled pairs for each domain and modality. 
Modality asymmetry alone, however, may produce accidental matches. For example, visually similar packaging does not necessarily indicate a meaningful graph relation. We therefore constrain bridge discovery with structural information in two stages. Candidate pairs are first restricted to PPR-based neighborhoods, limiting comparisons to nodes that are structurally relevant through the observed graph. For each remaining pair, we compute its PPR-neighborhood consistency as
\begin{equation}
r_{ij}
=
\frac{
\left|
\mathcal{N}_{\mathrm{PPR}}(v_i)
\cap
\mathcal{N}_{\mathrm{PPR}}(v_j)
\right|
}{
\min\left(
\left|\mathcal{N}_{\mathrm{PPR}}(v_i)\right|,
\left|\mathcal{N}_{\mathrm{PPR}}(v_j)\right|
\right)
}.
\end{equation}
A candidate pair is retained only if it satisfies both the modality-complementary criterion and a minimum PPR-overlap threshold. For each node, we keep at most $K_b$ retained pairs with the largest $r_{ij}$. The resulting sparse bridge set preserves modality-asymmetric associations that are also supported by the graph structure.

\subsection{Graph Context Construction}
The full augmented graph is too large to be serialized into the LLM input, while a conventional local subgraph remains dominated by fine-grained, domain-specific structural and multimodal patterns. Such a local view provides limited access to the shared high-level concepts needed for zero-shot transfer. CHARM therefore constructs a
compact target-specific context that combines local structural evidence with hierarchical semantic references and modality-complementary information.

Given a center node $v_c$, we first retrieve up to $K_p$ PPR-ranked neighbors from the original graph~\cite{pagerank,APPNP}, denoted as $\mathcal{N}_{\mathrm{PPR}}(v_c)$. These neighbors provide local structural evidence around the prediction target. We then use the precomputed synthetic hierarchy to retrieve the layer-1 anchor assigned to $v_c$, its corresponding layer-2 anchor, and the top-$m$ global anchors linked to the layer-2 anchor. This step adds semantic references at different abstraction levels, so that the LLM can reason over higher-level semantic concepts. Finally, for each center node, we keep up to $K_b$ bridge neighbors ranked by PPR-overlap score.

The resulting context is defined as
\begin{equation}
\mathcal{C}(v_c)=\{v_c\}\cup\mathcal{N}_{\mathrm{PPR}}(v_c)
\cup\mathcal{A}(v_c)\cup\mathcal{G}(v_c)\cup\mathcal{B}(v_c),
\end{equation}
where $\mathcal{A}(v_c)$ contains the retrieved layer-1 and layer-2 anchors, $\mathcal{G}(v_c)$ contains the top global anchors, and $\mathcal{B}(v_c)$ contains the selected bridge neighbors. Duplicated nodes are removed from the context. In this way, $\mathcal{C}(v_c)$ provides a compact view of the prediction target, combining local topology, semantic abstraction, and modality-specific supplementary information within a limited input budget. The bounded budget ensures that the input structure can be used across domains with different numbers of nodes.

\subsection{Graph Context Encoder}\label{sec:encoder}

The retrieved context brings together diverse forms of transferable
evidence. However, these context items vary in both modality reliability and structural role. First, text and image may contribute unequally across nodes and domains, making direct feature averaging unreliable. Second, serializing the context as a flat token sequence would obscure the relations among its items and require the frozen LLM to infer graph structure from token order alone. To address these two issues, \textsc{CHARM} introduces a graph context encoder with two successive stages. It first performs reliability-aware fusion for each context item, adaptively balancing its textual and visual evidence. It then propagates the fused representations over the context graph, enabling the retrieved items to exchange relational information before being converted into graph tokens.

For each context node $u\in\mathcal{C}(v_c)$, modality-specific
projectors first map its textual and visual features into the LLM
hidden space:
\begin{equation}
\mathbf{h}^{t}_u=\mathbf{W}_{t}\mathbf{x}^{t}_u,
\qquad
\mathbf{h}^{i}_u=\mathbf{W}_{i}\mathbf{x}^{i}_u.
\end{equation}
The two projectors are kept separate so that modality-specific
information is preserved before fusion. Since text and images may
provide unequal or even conflicting evidence, we construct an
interaction descriptor
\begin{equation}
\mathbf{q}_u=
[\mathbf{h}^{t}_u;\mathbf{h}^{i}_u;
\mathbf{h}^{t}_u\odot\mathbf{h}^{i}_u;
|\mathbf{h}^{t}_u-\mathbf{h}^{i}_u|].
\end{equation}
The first two terms retain the original modality-specific states, the
element-wise product captures dimension-wise agreement, and the
absolute difference reflects cross-modal discrepancy. Based on this interaction vector, a scalar modality gate is computed as
\begin{equation}
g_u=\sigma\!\left(\mathrm{MLP}_{g}(\mathbf{q}_u)\right),
\end{equation}
where $g_u\in[0,1]$ controls the relative contribution of the two
modalities. A larger value assigns more weight to textual evidence,
whereas a smaller value emphasizes the visual channel. The reweighted
states are then fused as
\begin{equation}
\mathbf{f}_u=
\mathrm{MLP}_{f}
\left(
[g_u\mathbf{h}^{t}_u;
(1-g_u)\mathbf{h}^{i}_u]
\right).
\end{equation}
The resulting representation $\mathbf{f}_u$ serves as its multimodal feature for subsequent context propagation. This node-wise gating preserves modality-specific cues without imposing a uniform fusion rule across heterogeneous contexts.

Reliability-aware fusion produces a multimodal state for each selected item, but it does not yet encode the relations among those items. We therefore construct a compact graph over
$\mathcal{C}(v_c)$. Let $\mathbf{A}^{\mathrm{ori}}_c$,
$\mathbf{A}^{\mathrm{hier}}_c$, and $\mathbf{A}^{\mathrm{br}}_c$
denote the context-restricted original, hierarchical, and bridge
adjacency matrices, respectively. Their union with self-loops is
defined as
\begin{equation}
\widetilde{\mathbf{A}}_c=
\left(
\mathbf{A}^{\mathrm{ori}}_c
\vee\mathbf{A}^{\mathrm{hier}}_c
\vee\mathbf{A}^{\mathrm{br}}_c
\right)
+\mathbf{I}.
\end{equation}
Original edges preserve co-purchase associations; hierarchical edges encode the semantic path $v\!\rightarrow\!a^{(1)}\!\rightarrow\!a^{(2)}\!\rightarrow\!p$; and bridge edges represent associations that are salient in one modality but weak in the other. The center node is connected to its retrieved evidence, ensuring that local neighbors, semantic references, and bridge items can directly contribute to the target representation.

After symmetric normalization,
\begin{equation}
\widehat{\mathbf{A}}_c=
\widetilde{\mathbf{D}}_c^{-1/2}
\widetilde{\mathbf{A}}_c
\widetilde{\mathbf{D}}_c^{-1/2},
\end{equation}
we apply two context propagation layers:
\begin{equation}
\mathbf{H}^{(\ell+1)}
=
\sigma\!\left(
\widehat{\mathbf{A}}_c
\mathbf{H}^{(\ell)}
\mathbf{W}^{(\ell)}
\right),
\end{equation}
where $\mathbf{H}^{(0)}$ stacks the initial representations of all
context nodes. Through propagation, the target node incorporates high-level semantic references and modality-complementary evidence from the retrieved context. The resulting graph tokens therefore provide the LLM with context-enriched representations, enabling it to make predictions based on the joint evidence rather than isolated node features. Although the context provides transferable semantics, messages from its items may obscure the fine-grained identity of the prediction target when propagated. We therefore introduce a
protection gate:
\begin{equation}
\mathbf{h}^{\mathrm{out}}_c
=
(1-\gamma)\mathbf{H}^{(0)}_c
+
\gamma\mathbf{H}^{(L)}_c,
\end{equation}
where $\gamma$ is learnable. The first term preserves the original
fused state and role information of the center, while the second
incorporates evidence propagated from the context. Non-center items
use their final propagated states. 

The final output is a sequence of continuous graph tokens:
\begin{equation}
\mathbf{S}_{v_c}=[\mathbf{s}_1,\ldots,\mathbf{s}_N]\in\mathbb{R}^{N\times d},
\end{equation}
where $d$ is the hidden dimension of the LLM. These tokens provide the interface between the constructed graph context and the LLM. The LLM receives the graph token sequence together with task instructions and performs zero-shot prediction without target-domain fine-tuning.

\begin{table*}[!t]
\centering
\caption{Zero-shot node classification results (\%).
The best and runner-up results for each metric are bolded and underlined,
respectively.}
\label{tab:nc}

\footnotesize
\setlength{\tabcolsep}{0.55pt}
\renewcommand{\arraystretch}{1.08}

\begin{tabular*}{\textwidth}
{@{\extracolsep{\fill}}lcccccccccc@{}}
\toprule
\multirow{2}{*}{Model}
& \multicolumn{2}{c}{\makecell{Grocery+Toys\\$\rightarrow$Movies}}
& \multicolumn{2}{c}{\makecell{Grocery+Movies\\$\rightarrow$Toys}}
& \multicolumn{2}{c}{\makecell{Toys+Movies\\$\rightarrow$Grocery}}
& \multicolumn{2}{c}{\makecell{Beauty+CD\\$\rightarrow$Arts}}
& \multicolumn{2}{c}{\makecell{Arts+Beauty\\$\rightarrow$CD}} \\
\cmidrule(lr){2-3}
\cmidrule(lr){4-5}
\cmidrule(lr){6-7}
\cmidrule(lr){8-9}
\cmidrule(l){10-11}
& Acc & F1 & Acc & F1 & Acc & F1 & Acc & F1 & Acc & F1 \\
\midrule

\multicolumn{11}{l}{\textbf{Conventional Supervised GNNs}} \\
GCN & 3.11 & 4.23 & 7.30 & 5.08 & 4.05 & 0.58 & 8.70 & 3.09 & 7.50 & 1.30 \\
GAT & 3.21 & 0.97 & 6.74 & 1.56 & 7.92 & 1.22 & 7.00 & 1.35 & 7.80 & 2.25 \\

\midrule
\multicolumn{11}{l}{\textbf{Graph Self-Supervised Methods}} \\
DGI & 2.25 & 0.70 & 3.67 & 1.21 & 8.64 & 1.46 & 14.30 & 3.10 & 8.80 & 2.19 \\
GraphMAE2 & \second{16.01} & 3.80 & 3.62 & 2.59 & 13.73 & 2.07 & 9.60 & 4.54 & 5.70 & 2.91 \\
BGRL & 2.67 & 1.86 & 4.40 & 3.53 & 8.23 & 4.29 & 7.30 & 3.41 & 4.00 & 2.25 \\

\midrule
\multicolumn{11}{l}{\textbf{GNN-Based Graph Foundation Models}} \\
UniGraph2 & 3.91 & 2.23 & 5.09 & 2.85 & 6.33 & 1.30 & 7.70 & 2.57 & 11.00 & 2.33 \\
SAMGPT & 5.32 & 2.18 & 5.11 & 4.44 & 6.14 & 1.81 & 9.70 & 4.95 & \second{11.20} & 5.46 \\
GCOPE & 4.98 & 2.83 & 6.21 & 4.42 & 2.55 & 1.33 & 9.00 & 4.81 & 6.40 & 1.86 \\

\midrule
\multicolumn{11}{l}{\textbf{LLM-Based Graph Foundation Models}} \\
GraphGPT & 10.07 & 4.22 & 4.42 & 4.79 & 6.00 & 4.91 & 5.83 & 3.52 & 5.24 & 3.08 \\
LLaGA & 10.07 & 0.62 & 2.27 & 0.19 & 11.36 & 0.56 & 6.43 & 4.49 & 10.45 & 3.13 \\
TEA-GLM & 6.62 & 2.50 & 7.80 & 3.90 & 5.12 & 3.10 & 7.63 & 6.08 & 4.66 & 2.56 \\
Graph4MM & 12.56 & \second{8.16} & 26.72 & \second{27.83} & 34.58 & \second{31.06} & 12.45 & \second{10.23} & 10.78 & 8.56 \\
MLaGA & 4.86 & 1.74 & \second{47.89} & 4.31 & \second{36.90} & 5.28 & \second{18.19} & 7.46 & 9.77 & \second{11.36} \\
\textsc{CHARM} & \best{17.10} & \best{16.45} & \best{50.33} & \best{48.72} & \best{59.12} & \best{49.77} & \best{22.78} & \best{10.31} & \best{22.60} & \best{17.34} \\

\bottomrule
\end{tabular*}
\end{table*} 

\subsection{Training and Inference}

For each source-domain configuration, \textsc{CHARM} jointly trains a single
model on node classification (NC) and link prediction (LP). Let
$\mathcal{D}_{\mathrm{src}}^{\mathrm{NC}}$ and
$\mathcal{D}_{\mathrm{src}}^{\mathrm{LP}}$ denote the corresponding
source-domain training sets. Each instance is paired with its task-specific
instruction and candidate answers, while the graph context encoder and
projection modules are shared across the two tasks. For NC, the LLM selects the
most appropriate category from the label descriptions; for LP, it determines
whether an edge exists between the queried nodes by choosing between
\emph{Yes} and \emph{No}.

The trainable graph modules are jointly optimized through the standard
next-token prediction loss of the frozen LLM:
\begin{equation}
\mathcal{L}
=
-\sum_{t\in\{\mathrm{NC},\mathrm{LP}\}}
\sum_{(x,y)\in\mathcal{D}_{\mathrm{src}}^{t}}
\log P_{\phi}
\left(
 y \mid \mathbf{S}_{x}(\theta), \mathbf{T}_{x}^{t}
\right),
\end{equation}
where $\mathbf{S}_{x}(\theta)$ denotes the graph-token sequence produced by
the trainable graph context encoder with parameters $\theta$,
$\mathbf{T}_{x}^{t}$ contains the task-specific instruction and candidate
answers, and $\phi$ denotes the fixed LLM parameters. For NC, $y$ is the
correct category option, whereas for LP,
$y\in\{\emph{Yes},\emph{No}\}$. During training, the text encoder, image
encoder, and LLM backbone remain frozen, and only the graph context encoder
and associated projection modules are updated. The same jointly trained
checkpoint is then directly evaluated on both tasks in the target domain
without parameter updates. The frozen LLM scores the fixed answer options at
the final token position, and the highest-scoring option is selected.
\section{Experiments}
\subsection{Baselines}
We compare four groups of methods. Conventional supervised GNNs include GCN and GAT~\cite{GCN,GAT}. Graph self-supervised methods include DGI, GraphMAE2, and BGRL~\cite{DGI,GraphMAE,GraphMAE2,BGRL,GraphSSLSurvey}. GNN-based GFMs include GCOPE, SAMGPT, and UniGraph2~\cite{GCOPE,SAMGPT,UniGraph2}. LLM-based GFMs include GraphGPT, LLaGA, TEA-GLM, Graph4MM, and MLaGA~\cite{GraphGPT,LLaGA,TEA-GLM,DBLP:conf/icml/NingFWXH25,DBLP:journals/corr/abs-2506-02568}.

Conventional supervised GNNs, graph self-supervised methods, and GNN-based GFMs are trained only on the source domains and then applied directly to the target graph. Text and image attributes are first encoded by CLIP and averaged to obtain the initial multimodal node features. For node classification, the graph encoder produces a target-node embedding, which is compared with the CLIP text embedding of each candidate label using cosine similarity; the label with the highest similarity is selected. Thus, these methods use the same open-vocabulary label set as CHARM, and no classifier is fitted on the target domain. For link prediction, the representations of the two queried endpoints are combined into a pair representation and fed into a binary linear classifier trained on positive and negative source-domain node pairs. All remaining hyperparameter settings follow their official implementations. LLM-based baselines are implemented using their publicly released code repositories and follow the corresponding default settings.

\begin{table*}[!t]
\centering
\caption{Zero-shot link prediction results (\%).
The best and runner-up results for each metric are bolded and underlined,
respectively.}
\label{tab:lp}

\footnotesize
\setlength{\tabcolsep}{0.55pt}
\renewcommand{\arraystretch}{1.08}

\begin{tabular*}{\textwidth}
{@{\extracolsep{\fill}}lcccccccccc@{}}
\toprule
\multirow{2}{*}{Model}
& \multicolumn{2}{c}{\makecell{Beauty+CD\\$\rightarrow$Arts}}
& \multicolumn{2}{c}{\makecell{Arts+Beauty\\$\rightarrow$CD}}
& \multicolumn{2}{c}{\makecell{Toys+Movies\\$\rightarrow$Grocery}}
& \multicolumn{2}{c}{\makecell{Grocery+Movies\\$\rightarrow$Toys}}
& \multicolumn{2}{c}{\makecell{Grocery+Toys\\$\rightarrow$Movies}} \\
\cmidrule(lr){2-3}
\cmidrule(lr){4-5}
\cmidrule(lr){6-7}
\cmidrule(lr){8-9}
\cmidrule(lr){10-11}
& Acc & F1
& Acc & F1
& Acc & F1
& Acc & F1
& Acc & F1 \\
\midrule

\multicolumn{11}{l}{\textbf{Conventional Supervised GNNs}} \\
GCN
& 53.02 & 42.75
& 50.76 & \best{58.05}
& 61.56 & 72.21
& 65.53 & 74.22
& 59.22 & 70.86 \\

GAT
& 54.38 & 59.25
& 51.84 & 57.41
& 60.88 & 70.23
& 62.72 & 70.91
& 67.94 & \second{74.53} \\

\midrule
\multicolumn{11}{l}{\textbf{Graph Self-Supervised Methods}} \\
DGI
& 49.61 & 45.96
& 50.83 & 21.38
& 58.94 & 70.33
& 63.28 & 72.40
& 64.44 & 66.98 \\

GraphMAE2
& 58.65 & \second{63.41}
& 50.97 & 54.29
& 51.38 & 67.28
& 67.50 & 56.63
& 64.44 & 72.38 \\

BGRL
& 52.10 & 58.35
& 50.30 & 45.03
& 56.56 & 69.66
& 55.03 & 68.98
& 60.63 & 71.71 \\

\midrule
\multicolumn{11}{l}{\textbf{GNN-Based Graph Foundation Models}} \\
UniGraph2
& 56.30 & 50.23
& 49.83 & 56.36
& 59.16 & 70.66
& 56.41 & 69.55
& 55.38 & 69.02 \\

SAMGPT
& 52.28 & 46.26
& 54.40 & 57.27
& 55.72 & 69.23
& 73.94 & 77.75
& \second{72.72} & \best{76.35} \\

GCOPE
& 51.89 & 49.73
& 52.18 & 54.62
& 63.34 & 71.95
& 50.00 & 0.00
& 50.00 & 0.00 \\

\midrule
\multicolumn{11}{l}{\textbf{LLM-Based Graph Foundation Models}} \\
GraphGPT
& \second{58.94} & 56.08
& \second{57.78} & 56.85
& 50.59 & 43.28
& 51.34 & 42.22
& 49.38 & 47.87 \\

LLaGA
& 48.59 & 35.42
& 49.78 & 36.15
& 55.25 & 67.69
& 50.38 & 65.40
& 51.81 & 64.17 \\

TEA-GLM
& 51.09 & 43.76
& 49.27 & 46.78
& 59.06 & 53.32
& 50.06 & 33.47
& 50.09 & 33.76 \\

Graph4MM
& 54.67 & 52.34
& 53.89 & 50.45
& 66.72 & 72.91
& 50.12 & 33.61
& 49.34 & 38.75 \\

MLaGA
& 55.22 & 49.91
& 55.69 & 39.51
& \second{72.81} & \second{74.89}
& \second{76.91} & \second{80.72}
& 68.70 & 68.31 \\

\textsc{CHARM}
& \best{59.28} & \best{65.85}
& \best{58.13} & \second{57.52}
& \best{73.41} & \best{75.63}
& \best{84.26} & \best{84.04}
& \best{74.31} & 72.30 \\

\bottomrule
\end{tabular*}
\end{table*}

\subsection{Experimental Setting}
We evaluate on six multimodal Amazon product graphs. Grocery, Movies, and Toys are adopted from the MAGB graph benchmark~\cite{DBLP:conf/kdd/0004L0YHLZ0W25}. We further construct Arts, Beauty, and CD by filtering the original Amazon product data~\cite{DataAmazon} to retain products with valid textual and visual attributes. The resulting Arts, Beauty, and CD graphs contain 11,640, 11,874, and 11,851 nodes, with 13, 9, and 19 classes, respectively. Edges represent co-purchase relations. GMT evaluates each domain once as target; ABC evaluates Arts and CD, with Beauty used only as source. For example, Grocery+Movies$\rightarrow$Toys indicates
that the model is trained on Grocery and Movies and directly evaluated on Toys. The target graph topology and attributes are used only to construct unlabeled graph contexts.

For each transfer setting, \textsc{CHARM} is trained once by jointly using the
NC and LP examples from the two source domains. The two tasks use separate
prompt templates and answer spaces but share all trainable graph modules. The
resulting single checkpoint is evaluated on both NC and LP in the held-out
target domain, rather than training a separate model for each task.

For node classification (NC), the candidate category names of the target dataset are provided as answer options. The model computes the option score of each category and selects the highest-scoring one. We report Accuracy and Macro-F1. For link prediction (LP), positive samples are observed edges, whereas negative samples are randomly drawn from nonadjacent node pairs. Each graph contains 10,000 source-domain training pairs and 3,200 target-domain test pairs, with equal numbers of positive and negative samples. For a positive test pair, the queried edge is removed before PPR computation, bridge retrieval, and context construction.

\subsection{Zero-Shot Node Classification}

Table~\ref{tab:nc} reports the zero-shot node classification results on five target domains. Overall, CHARM achieves the highest Accuracy in every transfer setting, demonstrating consistent generalization across both dataset groups. On the three GMT transfers, CHARM obtains accuracies of 17.10\%, 50.33\%, and 59.12\% on Movies, Toys, and Grocery, respectively, together with the best Macro-F1 scores. The simultaneous gains in Accuracy and Macro-F1 indicate that the improvements are not obtained merely by favoring a small number of frequent categories. The gap between the two metrics also reflects the effect of class imbalance. For example, MLaGA obtains relatively high Accuracy but much lower Macro-F1 on Toys and Grocery. This pattern is consistent with strong performance on a few frequent head classes but limited coverage of tail classes. Because Macro-F1 assigns equal weight to every class, it penalizes such class-wise imbalance more strongly than Accuracy. CHARM also achieves the highest Accuracy on Arts and CD, where the source and target domains exhibit larger differences in vocabulary and visual appearance. This result shows that its transferability is not limited to the relatively related domains within the GMT group.

A comparison across baselines provides further insight into these results. Conventional supervised GNNs and graph self-supervised methods learn structural representations from the source domains, but their generally limited performance suggests that directly matching graph embeddings with CLIP label embeddings is insufficient for zero-shot open-vocabulary classification. Even when the graph encoder produces informative source-domain representations, these representations may not be well aligned with the semantic space of unseen category names. GraphGPT, LLaGA, and TEA-GLM improve the graph--language interface through graph--text grounding, structure-aware node sequences, or token-level alignment. However, their performance varies substantially across target domains, possibly because they are primarily developed for text-attributed graphs and provide limited modeling of how textual and visual evidence jointly supports category prediction. Graph4MM and MLaGA achieve stronger results in several settings by incorporating multimodal attributes and graph structure, confirming the importance of multimodal information. Nevertheless, their inconsistent performance across targets indicates that the learned alignment may remain sensitive to domain-specific vocabulary, visual appearance, and local relation patterns.

In comparison, CHARM maintains strong performance across target domains with different category spaces and multimodal distributions. This consistency is particularly important in the zero-shot setting, where no target labels are available to train a classifier or adapt the model. The results suggest that CHARM produces contextualized representations that retain instance-level discriminative information while exposing higher-level semantics that can be associated with candidate category names. Consequently, the frozen LLM can infer the category of an unseen target node from its multimodal and relational evidence.

\subsection{Zero-Shot Link Prediction}

Table~\ref{tab:lp} reports zero-shot LP results on five target domains. CHARM achieves the highest Accuracy in every setting, improving over the strongest baseline by 0.34, 0.35, 0.60, 7.35, and 1.59 percentage points on Arts, CD, Grocery, Toys, and Movies, respectively. It obtains the best F1 on Arts, Grocery, and Toys, ranks second on CD, and remains competitive on Movies. On Movies, the best Accuracy but lower F1 suggests that residual errors are concentrated in one class, revealing a remaining calibration challenge under domain shift.

The comparison across baseline families highlights the difficulty of zero-shot LP. Conventional and self-supervised GNNs transfer local structural patterns inconsistently across domains. GNN-based graph foundation models improve cross-graph transfer but usually predict from independently encoded endpoints, limiting relational-context modeling. LLM-based methods benefit from graph--language or multimodal alignment: GraphGPT is strong on Arts and CD, whereas MLaGA performs well on Grocery and Toys. Their variation across targets nevertheless indicates sensitivity to domain-specific edge semantics. In contrast, CHARM evaluates a queried pair through the structured multimodal contexts of both endpoints rather than isolated representations. Integrating local structure, multi-level semantic references, and modality-complementary evidence enables the frozen LLM to assess endpoint compatibility under the target-domain context, reducing dependence on source-specific edge patterns.

\begin{table}[t]
\centering
\caption{Ablation results on zero-shot NC (\%).}
\label{tab:ablation}

\small
\setlength{\tabcolsep}{1.8pt}
\renewcommand{\arraystretch}{1.02}

\begin{tabular*}{\columnwidth}
{@{\extracolsep{\fill}}lcccc@{}}
\toprule
\multirow{2}{*}{Variant}
& \multicolumn{2}{c}{T+M$\rightarrow$G}
& \multicolumn{2}{c}{G+M$\rightarrow$T} \\
\cmidrule(lr){2-3}
\cmidrule(l){4-5}
& Acc & F1 & Acc & F1 \\
\midrule

Full Model
& \best{59.12} & \best{49.77}
& \best{50.33} & \best{48.72} \\

\midrule

w/o Layer-1
& 36.84 & 34.28
& \second{50.21} & 43.23 \\

w/o Layer-2
& 53.47 & 45.36
& 33.44 & 29.55 \\

w/o Abstract Nodes
& 35.83 & 33.97
& 38.58 & 37.22 \\

Center Only
& 3.16 & 1.38
& 7.25 & 1.96 \\

w/o PPR
& 47.38 & 39.74
& 43.39 & 38.94 \\

w/o Bridge
& 47.00 & 39.09
& 39.67 & 36.30 \\

w/o Fusion
& 33.46 & 27.19
& 49.96 & \second{48.35} \\

w/o Context Prop.
& 45.94 & 42.02
& 39.18 & 32.25 \\

w/o Center Protection
& \second{53.50} & \second{47.38}
& 47.09 & 45.71 \\

\bottomrule
\end{tabular*}
\end{table}

\subsection{Model Analysis}\label{sec:model_analysis}

Table~\ref{tab:ablation} first evaluates the hierarchical semantic components that support cross-domain transfer. Removing all abstract nodes decreases Accuracy by 23.29 and 11.75 percentage points on Grocery and Toys, respectively, showing that raw target nodes and their local neighborhoods alone do not provide sufficiently transferable semantic references. The layer-wise ablations further demonstrate that different abstraction levels play complementary roles: removing Layer-1 anchors causes a substantial degradation on Grocery but only a minor change on Toys, whereas removing Layer-2 anchors has a much stronger effect on Toys. This difference suggests that Layer-1 anchors preserve relatively fine-grained semantic groups close to individual instances, while Layer-2 anchors capture broader in-domain concepts. Retaining both levels therefore allows \textsc{CHARM} to preserve instance-relevant distinctions while progressively connecting them to broader concepts and global anchors. The near-chance performance of the Center Only variant further confirms that the center-node representation alone is insufficient and that effective zero-shot transfer requires both neighborhood evidence and hierarchical semantic references.

\begin{figure}[t]
\centering
\includegraphics[width=0.85\columnwidth]{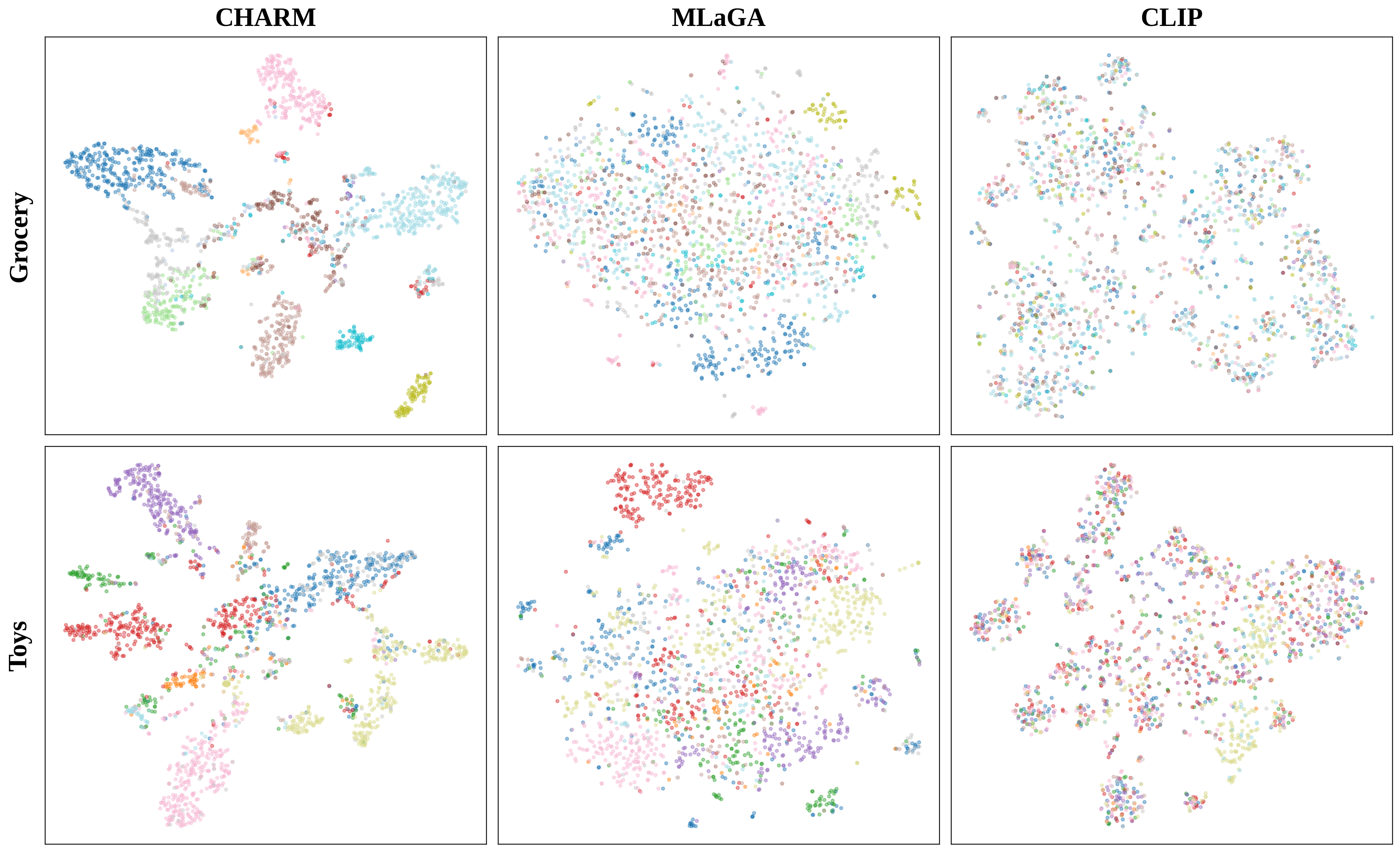}
\caption{t-SNE representations on Grocery and Toys.}
\label{fig:tsne}
\Description{Six t-SNE plots compare CHARM, MLaGA, and CLIP representations on Grocery and Toys. CHARM shows more compact class clusters and clearer separation.}
\end{figure}
Removing PPR neighbors or modality-complementary bridges consistently reduces performance, demonstrating the importance of both local structural information and relations supported strongly by either text or image. The learned fusion module has a particularly large effect on Grocery but a smaller effect on Toys, suggesting that the relative reliability of textual and visual evidence varies across domains. Context propagation is also important: removing it decreases Accuracy by more than 11 points on both targets, showing that simply retrieving relevant context items is insufficient and that their structural and semantic relations must be integrated before prediction. Center protection provides smaller but consistent gains by retaining the original identity of the target node while incorporating contextual information.

\begin{wrapfigure}[14]{r}{0.48\columnwidth}
\vspace{-6pt}
\centering
\includegraphics[width=0.96\linewidth]{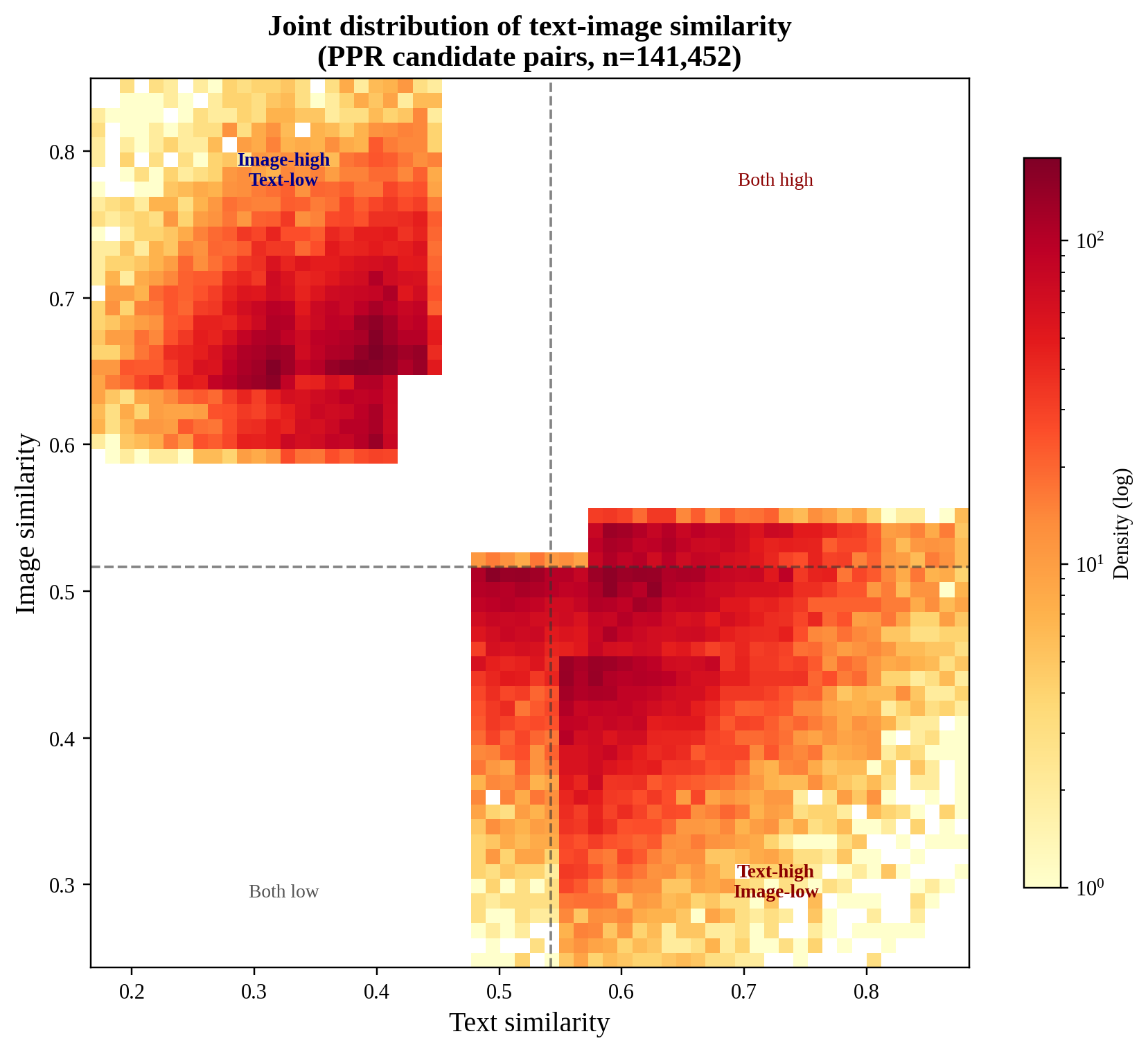}
\vspace{-5pt}
\caption{Text--image similarity of PPR pairs.}
\label{fig:bridge_analysis}
\Description{A density plot shows that many structurally related PPR candidate pairs are strong in only text or only image similarity.}
\vspace{-6pt}
\end{wrapfigure}

Figure~\ref{fig:bridge_analysis} provides further evidence for modality-complementary modeling, showing that many PPR-restricted pairs occupy asymmetric text--image regions rather than the high--high region and that structurally relevant relations are often evident in only one modality. An agreement-only criterion would discard these pairs, while fixed equal-weight fusion could overemphasize the weaker modality; the complementary bridges and learned fusion mechanism instead preserve such asymmetric evidence and adapt its contribution to each context item. Figure~\ref{fig:tsne} presents the corresponding representation-level comparison: frozen CLIP features remain substantially mixed, while MLaGA only partially organizes the samples into class-consistent regions, whereas \textsc{CHARM} produces more compact groups and clearer separation on both targets. These observations are consistent with the ablation results: hierarchical references reduce dependence on domain-specific details, while reliability-aware fusion and relation-aware propagation organize multimodal and structural evidence.
\section{Conclusion}
We presented \textsc{CHARM}, advancing zero-shot transfer on multimodal graphs by treating structured graph contexts as transferable units rather than isolated nodes. Its hierarchical semantic modeling and modality-aware context encoding jointly preserve cross-domain abstraction, instance-level distinctions, and complementary multimodal evidence. Results across five target domains validate its effectiveness without target labels or backbone adaptation.

\label{content:end}

\clearpage
\bibliographystyle{ACM-Reference-Format}
\bibliography{references}

@inproceedings{Social_nets,
  author       = {Jie Tang and
                  Jimeng Sun and
                  Chi Wang and
                  Zi Yang},
  editor       = {John F. Elder IV and
                  Fran{\c{c}}oise Fogelman{-}Souli{\'{e}} and
                  Peter A. Flach and
                  Mohammed Javeed Zaki},
  title        = {Social influence analysis in large-scale networks},
  booktitle    = {Proceedings of the 15th {ACM} {SIGKDD} International Conference on
                  Knowledge Discovery and Data Mining, Paris, France, June 28 - July
                  1, 2009},
  pages        = {807--816},
  publisher    = {{ACM}},
  year         = {2009},
  url          = {https://doi.org/10.1145/1557019.1557108},
  doi          = {10.1145/1557019.1557108},
  bibsource    = {dblp computer science bibliography, https://dblp.org}
}

@inproceedings{Message-passing,
  title={Neural message passing for quantum chemistry},
  author={Gilmer, Justin and Schoenholz, Samuel S and Riley, Patrick F and Vinyals, Oriol and Dahl, George E},
  booktitle={International conference on machine learning},
  pages={1263--1272},
  year={2017},
  organization={Pmlr}
}

@article{Survey-GNNS,
  author       = {Zonghan Wu and
                  Shirui Pan and
                  Fengwen Chen and
                  Guodong Long and
                  Chengqi Zhang and
                  Philip S. Yu},
  title        = {A Comprehensive Survey on Graph Neural Networks},
  journal      = {{IEEE} Trans. Neural Networks Learn. Syst.},
  volume       = {32},
  number       = {1},
  pages        = {4--24},
  year         = {2021},
  url          = {https://doi.org/10.1109/TNNLS.2020.2978386},
  doi          = {10.1109/TNNLS.2020.2978386},
  bibsource    = {dblp computer science bibliography, https://dblp.org}
}

@inproceedings{GAT,
  author       = {Petar Velickovic and
                  Guillem Cucurull and
                  Arantxa Casanova and
                  Adriana Romero and
                  Pietro Li{\`{o}} and
                  Yoshua Bengio},
  title        = {Graph Attention Networks},
  booktitle    = {6th International Conference on Learning Representations, {ICLR} 2018,
                  Vancouver, BC, Canada, April 30 - May 3, 2018, Conference Track Proceedings},
  publisher    = {OpenReview.net},
  year         = {2018},
  url          = {https://openreview.net/forum?id=rJXMpikCZ},
  bibsource    = {dblp computer science bibliography, https://dblp.org}
}

@inproceedings{GraphSAGE,
  author       = {William L. Hamilton and
                  Zhitao Ying and
                  Jure Leskovec},
  editor       = {Isabelle Guyon and
                  Ulrike von Luxburg and
                  Samy Bengio and
                  Hanna M. Wallach and
                  Rob Fergus and
                  S. V. N. Vishwanathan and
                  Roman Garnett},
  title        = {Inductive Representation Learning on Large Graphs},
  booktitle    = {Advances in Neural Information Processing Systems 30: Annual Conference
                  on Neural Information Processing Systems 2017, December 4-9, 2017,
                  Long Beach, CA, {USA}},
  pages        = {1024--1034},
  year         = {2017},
  url          = {https://proceedings.neurips.cc/paper/2017/hash/5dd9db5e033da9c6fb5ba83c7a7ebea9-Abstract.html},
  bibsource    = {dblp computer science bibliography, https://dblp.org}
}

@inproceedings{GIN,
  author       = {Keyulu Xu and
                  Weihua Hu and
                  Jure Leskovec and
                  Stefanie Jegelka},
  title        = {How Powerful are Graph Neural Networks?},
  booktitle    = {7th International Conference on Learning Representations, {ICLR} 2019,
                  New Orleans, LA, USA, May 6-9, 2019},
  publisher    = {OpenReview.net},
  year         = {2019},
  url          = {https://openreview.net/forum?id=ryGs6iA5Km},
  bibsource    = {dblp computer science bibliography, https://dblp.org}
}

@techreport{pagerank,
  title={The PageRank citation ranking: Bringing order to the web.},
  author={Page, Lawrence and Brin, Sergey and Motwani, Rajeev and Winograd, Terry},
  year={1999},
  institution={Stanford infolab}
}

@inproceedings{APPNP,
  author       = {Johannes Klicpera and
                  Aleksandar Bojchevski and
                  Stephan G{\"{u}}nnemann},
  title        = {Predict then Propagate: Graph Neural Networks meet Personalized PageRank},
  booktitle    = {7th International Conference on Learning Representations, {ICLR} 2019,
                  New Orleans, LA, USA, May 6-9, 2019},
  publisher    = {OpenReview.net},
  year         = {2019},
}

@inproceedings{Transformer,
  author       = {Ashish Vaswani and
                  Noam Shazeer and
                  Niki Parmar and
                  Jakob Uszkoreit and
                  Llion Jones and
                  Aidan N. Gomez and
                  Lukasz Kaiser and
                  Illia Polosukhin},
  editor       = {Isabelle Guyon and
                  Ulrike von Luxburg and
                  Samy Bengio and
                  Hanna M. Wallach and
                  Rob Fergus and
                  S. V. N. Vishwanathan and
                  Roman Garnett},
  title        = {Attention is All you Need},
  booktitle    = {Advances in Neural Information Processing Systems 30: Annual Conference
                  on Neural Information Processing Systems 2017, December 4-9, 2017,
                  Long Beach, CA, {USA}},
  pages        = {5998--6008},
  year         = {2017},
  url          = {https://proceedings.neurips.cc/paper/2017/hash/3f5ee243547dee91fbd053c1c4a845aa-Abstract.html},
  bibsource    = {dblp computer science bibliography, https://dblp.org}
}

@inproceedings{RiemannGFM,
  author       = {Li Sun and
                  Zhenhao Huang and
                  Suyang Zhou and
                  Qiqi Wan and
                  Hao Peng and
                  Philip S. Yu},
  editor       = {Guodong Long and
                  Michale Blumestein and
                  Yi Chang and
                  Liane Lewin{-}Eytan and
                  Zi Helen Huang and
                  Elad Yom{-}Tov},
  title        = {RiemannGFM: Learning a Graph Foundation Model from Riemannian Geometry},
  booktitle    = {Proceedings of the {ACM} on Web Conference 2025, {WWW} 2025, Sydney,
                  NSW, Australia, 28 April 2025- 2 May 2025},
  pages        = {1154--1165},
  publisher    = {{ACM}},
  year         = {2025},
  url          = {https://doi.org/10.1145/3696410.3714952},
  doi          = {10.1145/3696410.3714952},
  bibsource    = {dblp computer science bibliography, https://dblp.org}
}

@article{AnyGraph,
  author       = {Lianghao Xia and
                  Chao Huang},
  title        = {AnyGraph: Graph Foundation Model in the Wild},
  journal      = {CoRR},
  volume       = {abs/2408.10700},
  year         = {2024},
  url          = {https://doi.org/10.48550/arXiv.2408.10700},
  doi          = {10.48550/ARXIV.2408.10700},
  eprinttype    = {arXiv},
  eprint       = {2408.10700},
  bibsource    = {dblp computer science bibliography, https://dblp.org}
}

@inproceedings{DGI,
  author       = {Petar Velickovic and
                  William Fedus and
                  William L. Hamilton and
                  Pietro Li{\`{o}} and
                  Yoshua Bengio and
                  R. Devon Hjelm},
  title        = {Deep Graph Infomax},
  booktitle    = {7th International Conference on Learning Representations, {ICLR} 2019,
                  New Orleans, LA, USA, May 6-9, 2019},
  publisher    = {OpenReview.net},
  year         = {2019},
  url          = {https://openreview.net/forum?id=rklz9iAcKQ},
  bibsource    = {dblp computer science bibliography, https://dblp.org}
}

@inproceedings{BGRL,
  title={Bootstrapped representation learning on graphs},
  author={Thakoor, Shantanu and Tallec, Corentin and Azar, Mohammad Gheshlaghi and Munos, R{\'e}mi and Veli{\v{c}}kovi{\'c}, Petar and Valko, Michal},
  booktitle={ICLR 2021 Workshop on Geometrical and Topological Representation Learning},
  year={2021}
}

@inproceedings{GCN,
  author       = {Thomas N. Kipf and
                  Max Welling},
  title        = {Semi-Supervised Classification with Graph Convolutional Networks},
  booktitle    = {5th International Conference on Learning Representations, {ICLR} 2017,
                  Toulon, France, April 24-26, 2017, Conference Track Proceedings},
  publisher    = {OpenReview.net},
  year         = {2017},
  url          = {https://openreview.net/forum?id=SJU4ayYgl},
  bibsource    = {dblp computer science bibliography, https://dblp.org}
}

@inproceedings{DataAmazon,
  author       = {Julian J. McAuley and
                  Christopher Targett and
                  Qinfeng Shi and
                  Anton van den Hengel},
  title        = {Image-Based Recommendations on Styles and Substitutes},
  booktitle    = {Proceedings of the 38th International {ACM} {SIGIR} Conference on
                  Research and Development in Information Retrieval, Santiago, Chile,
                  August 9-13, 2015},
  pages        = {43--52},
  publisher    = {{ACM}},
  year         = {2015},
  url          = {https://doi.org/10.1145/2766462.2767755},
  bibsource    = {dblp computer science bibliography, https://dblp.org}
}

@inproceedings{GraphMAE,
  author       = {Zhenyu Hou and
                  Xiao Liu and
                  Yukuo Cen and
                  Yuxiao Dong and
                  Hongxia Yang and
                  Chunjie Wang and
                  Jie Tang},
  title        = {GraphMAE: Self-Supervised Masked Graph Autoencoders},
  booktitle    = {{KDD} '22: The 28th {ACM} {SIGKDD} Conference on Knowledge Discovery
                  and Data Mining, Washington, DC, USA, August 14 - 18, 2022},
  pages        = {594--604},
  publisher    = {{ACM}},
  year         = {2022},
  url          = {https://doi.org/10.1145/3534678.3539321},
  bibsource    = {dblp computer science bibliography, https://dblp.org}
}

@inproceedings{GraphMAE2,
  author       = {Zhenyu Hou and
                  Yufei He and
                  Yukuo Cen and
                  Xiao Liu and
                  Yuxiao Dong and
                  Evgeny Kharlamov and
                  Jie Tang},
  editor       = {Ying Ding and
                  Jie Tang and
                  Juan F. Sequeda and
                  Lora Aroyo and
                  Carlos Castillo and
                  Geert{-}Jan Houben},
  title        = {GraphMAE2: {A} Decoding-Enhanced Masked Self-Supervised Graph Learner},
  booktitle    = {Proceedings of the {ACM} Web Conference 2023, {WWW} 2023, Austin,
                  TX, USA, 30 April 2023 - 4 May 2023},
  pages        = {737--746},
  publisher    = {{ACM}},
  year         = {2023},
  url          = {https://doi.org/10.1145/3543507.3583379},
  doi          = {10.1145/3543507.3583379},
  bibsource    = {dblp computer science bibliography, https://dblp.org}
}

@inproceedings{OFA,
  author       = {Hao Liu and
                  Jiarui Feng and
                  Lecheng Kong and
                  Ningyue Liang and
                  Dacheng Tao and
                  Yixin Chen and
                  Muhan Zhang},
  title        = {One For All: Towards Training One Graph Model For All Classification
                  Tasks},
  booktitle    = {The Twelfth International Conference on Learning Representations,
                  {ICLR} 2024, Vienna, Austria, May 7-11, 2024},
  publisher    = {OpenReview.net},
  year         = {2024},
  url          = {https://openreview.net/forum?id=4IT2pgc9v6},
  bibsource    = {dblp computer science bibliography, https://dblp.org}
}

@inproceedings{UniGraph2,
  author       = {Yufei He and
                  Yuan Sui and
                  Xiaoxin He and
                  Yue Liu and
                  Yifei Sun and
                  Bryan Hooi},
  editor       = {Guodong Long and
                  Michale Blumestein and
                  Yi Chang and
                  Liane Lewin{-}Eytan and
                  Zi Helen Huang and
                  Elad Yom{-}Tov},
  title        = {UniGraph2: Learning a Unified Embedding Space to Bind Multimodal Graphs},
  booktitle    = {Proceedings of the {ACM} on Web Conference 2025, {WWW} 2025, Sydney,
                  NSW, Australia, 28 April 2025- 2 May 2025},
  pages        = {1759--1770},
  publisher    = {{ACM}},
  year         = {2025},
  url          = {https://doi.org/10.1145/3696410.3714818},
  doi          = {10.1145/3696410.3714818},
  bibsource    = {dblp computer science bibliography, https://dblp.org}
}

@inproceedings{ OpenGraph,
  author       = {Lianghao Xia and
                  Ben Kao and
                  Chao Huang},
  editor       = {Yaser Al{-}Onaizan and
                  Mohit Bansal and
                  Yun{-}Nung Chen},
  title        = {OpenGraph: Towards Open Graph Foundation Models},
  booktitle    = {Findings of the Association for Computational Linguistics: {EMNLP}
                  2024, Miami, Florida, USA, November 12-16, 2024},
  pages        = {2365--2379},
  publisher    = {Association for Computational Linguistics},
  year         = {2024},
  url          = {https://doi.org/10.18653/v1/2024.findings-emnlp.132},
  doi          = {10.18653/V1/2024.FINDINGS-EMNLP.132},
  bibsource    = {dblp computer science bibliography, https://dblp.org}
}

@inproceedings{ SAMGPT,
  author       = {Xingtong Yu and
                  Zechuan Gong and
                  Chang Zhou and
                  Yuan Fang and
                  Hui Zhang},
  editor       = {Guodong Long and
                  Michale Blumestein and
                  Yi Chang and
                  Liane Lewin{-}Eytan and
                  Zi Helen Huang and
                  Elad Yom{-}Tov},
  title        = {{SAMGPT:} Text-free Graph Foundation Model for Multi-domain Pre-training
                  and Cross-domain Adaptation},
  booktitle    = {Proceedings of the {ACM} on Web Conference 2025, {WWW} 2025, Sydney,
                  NSW, Australia, 28 April 2025- 2 May 2025},
  pages        = {1142--1153},
  publisher    = {{ACM}},
  year         = {2025},
  url          = {https://doi.org/10.1145/3696410.3714828},
  doi          = {10.1145/3696410.3714828},
  bibsource    = {dblp computer science bibliography, https://dblp.org}
}

@inproceedings{ TEA-GLM,
  author       = {Duo Wang and
                  Yuan Zuo and
                  Fengzhi Li and
                  Junjie Wu},
  editor       = {Amir Globersons and
                  Lester Mackey and
                  Danielle Belgrave and
                  Angela Fan and
                  Ulrich Paquet and
                  Jakub M. Tomczak and
                  Cheng Zhang},
  title        = {LLMs as Zero-shot Graph Learners: Alignment of {GNN} Representations
                  with {LLM} Token Embeddings},
  booktitle    = {Advances in Neural Information Processing Systems 38: Annual Conference
                  on Neural Information Processing Systems 2024, NeurIPS 2024, Vancouver,
                  BC, Canada, December 10 - 15, 2024},
  year         = {2024},
  url          = {http://papers.nips.cc/paper\_files/paper/2024/hash/0b77d3a82b59e9d9899370b378087faf-Abstract-Conference.html},
  bibsource    = {dblp computer science bibliography, https://dblp.org}
}

@article{PFMs,
  title={Graph Foundation Models: A Comprehensive Survey},
  author={Wang, Zehong and Liu, Zheyuan and Ma, Tianyi and Li, Jiazheng and Zhang, Zheyuan and Fu, Xingbo and Li, Yiyang and Yuan, Zhengqing and Song, Wei and Ma, Yijun and others},
  journal={arXiv preprint arXiv:2505.15116},
  year={2025}
}

@inproceedings{GraphCLIP,
  author       = {Yun Zhu and
                  Haizhou Shi and
                  Xiaotang Wang and
                  Yongchao Liu and
                  Yaoke Wang and
                  Boci Peng and
                  Chuntao Hong and
                  Siliang Tang},
  editor       = {Guodong Long and
                  Michale Blumestein and
                  Yi Chang and
                  Liane Lewin{-}Eytan and
                  Zi Helen Huang and
                  Elad Yom{-}Tov},
  title        = {GraphCLIP: Enhancing Transferability in Graph Foundation Models for
                  Text-Attributed Graphs},
  booktitle    = {Proceedings of the {ACM} on Web Conference 2025, {WWW} 2025, Sydney,
                  NSW, Australia, 28 April 2025- 2 May 2025},
  pages        = {2183--2197},
  publisher    = {{ACM}},
  year         = {2025},
  url          = {https://doi.org/10.1145/3696410.3714801},
  doi          = {10.1145/3696410.3714801},
  bibsource    = {dblp computer science bibliography, https://dblp.org}
}

@inproceedings{HiGPT,
  author       = {Jiabin Tang and
                  Yuhao Yang and
                  Wei Wei and
                  Lei Shi and
                  Long Xia and
                  Dawei Yin and
                  Chao Huang},
  editor       = {Ricardo Baeza{-}Yates and
                  Francesco Bonchi},
  title        = {HiGPT: Heterogeneous Graph Language Model},
  booktitle    = {Proceedings of the 30th {ACM} {SIGKDD} Conference on Knowledge Discovery
                  and Data Mining, {KDD} 2024, Barcelona, Spain, August 25-29, 2024},
  pages        = {2842--2853},
  publisher    = {{ACM}},
  year         = {2024},
  url          = {https://doi.org/10.1145/3637528.3671987},
  doi          = {10.1145/3637528.3671987},
  bibsource    = {dblp computer science bibliography, https://dblp.org}
}

@inproceedings{GCOPE,
  author       = {Haihong Zhao and
                  Aochuan Chen and
                  Xiangguo Sun and
                  Hong Cheng and
                  Jia Li},
  editor       = {Ricardo Baeza{-}Yates and
                  Francesco Bonchi},
  title        = {All in One and One for All: {A} Simple yet Effective Method towards
                  Cross-domain Graph Pretraining},
  booktitle    = {Proceedings of the 30th {ACM} {SIGKDD} Conference on Knowledge Discovery
                  and Data Mining, {KDD} 2024, Barcelona, Spain, August 25-29, 2024},
  pages        = {4443--4454},
  publisher    = {{ACM}},
  year         = {2024},
  url          = {https://doi.org/10.1145/3637528.3671913},
  doi          = {10.1145/3637528.3671913},
  bibsource    = {dblp computer science bibliography, https://dblp.org}
}

@inproceedings{GOFA,
  author       = {Lecheng Kong and
                  Jiarui Feng and
                  Hao Liu and
                  Chengsong Huang and
                  Jiaxin Huang and
                  Yixin Chen and
                  Muhan Zhang},
  title        = {{GOFA:} {A} Generative One-For-All Model for Joint Graph Language
                  Modeling},
  booktitle    = {The Thirteenth International Conference on Learning Representations,
                  {ICLR} 2025, Singapore, April 24-28, 2025},
  publisher    = {OpenReview.net},
  year         = {2025},
  url          = {https://openreview.net/forum?id=mIjblC9hfm},
  bibsource    = {dblp computer science bibliography, https://dblp.org}
}

@article{MDGPT,
  author       = {Xingtong Yu and
                  Chang Zhou and
                  Yuan Fang and
                  Xinming Zhang},
  title        = {Text-Free Multi-domain Graph Pre-training: Toward Graph Foundation
                  Models},
  journal      = {CoRR},
  volume       = {abs/2405.13934},
  year         = {2024},
  url          = {https://doi.org/10.48550/arXiv.2405.13934},
  doi          = {10.48550/ARXIV.2405.13934},
  eprinttype    = {arXiv},
  eprint       = {2405.13934},
  bibsource    = {dblp computer science bibliography, https://dblp.org}
}

@inproceedings{ZeroG,
  author       = {Yuhan Li and
                  Peisong Wang and
                  Zhixun Li and
                  Jeffrey Xu Yu and
                  Jia Li},
  editor       = {Ricardo Baeza{-}Yates and
                  Francesco Bonchi},
  title        = {ZeroG: Investigating Cross-dataset Zero-shot Transferability in Graphs},
  booktitle    = {Proceedings of the 30th {ACM} {SIGKDD} Conference on Knowledge Discovery
                  and Data Mining, {KDD} 2024, Barcelona, Spain, August 25-29, 2024},
  pages        = {1725--1735},
  publisher    = {{ACM}},
  year         = {2024},
  url          = {https://doi.org/10.1145/3637528.3671982},
  doi          = {10.1145/3637528.3671982},
  bibsource    = {dblp computer science bibliography, https://dblp.org}
}

@inproceedings{UniGraph,
  author       = {Yufei He and
                  Yuan Sui and
                  Xiaoxin He and
                  Bryan Hooi},
  editor       = {Yizhou Sun and
                  Flavio Chierichetti and
                  Hady W. Lauw and
                  Claudia Perlich and
                  Wee Hyong Tok and
                  Andrew Tomkins},
  title        = {UniGraph: Learning a Unified Cross-Domain Foundation Model for Text-Attributed
                  Graphs},
  booktitle    = {Proceedings of the 31st {ACM} {SIGKDD} Conference on Knowledge Discovery
                  and Data Mining, V.1, {KDD} 2025, Toronto, ON, Canada, August 3-7,
                  2025},
  pages        = {448--459},
  publisher    = {{ACM}},
  year         = {2025},
  url          = {https://doi.org/10.1145/3690624.3709277},
  doi          = {10.1145/3690624.3709277},
  bibsource    = {dblp computer science bibliography, https://dblp.org}
}

@inproceedings{LLaGA,
  author       = {Runjin Chen and
                  Tong Zhao and
                  Ajay Kumar Jaiswal and
                  Neil Shah and
                  Zhangyang Wang},
  title        = {LLaGA: Large Language and Graph Assistant},
  booktitle    = {Forty-first International Conference on Machine Learning, {ICML} 2024,
                  Vienna, Austria, July 21-27, 2024},
  publisher    = {OpenReview.net},
  year         = {2024},
  url          = {https://openreview.net/forum?id=B48Pzc4oKi},
  bibsource    = {dblp computer science bibliography, https://dblp.org}
}

@inproceedings{GraphGPT,
  author       = {Jiabin Tang and
                  Yuhao Yang and
                  Wei Wei and
                  Lei Shi and
                  Lixin Su and
                  Suqi Cheng and
                  Dawei Yin and
                  Chao Huang},
  editor       = {Grace Hui Yang and
                  Hongning Wang and
                  Sam Han and
                  Claudia Hauff and
                  Guido Zuccon and
                  Yi Zhang},
  title        = {GraphGPT: Graph Instruction Tuning for Large Language Models},
  booktitle    = {Proceedings of the 47th International {ACM} {SIGIR} Conference on
                  Research and Development in Information Retrieval, {SIGIR} 2024, Washington
                  DC, USA, July 14-18, 2024},
  pages        = {491--500},
  publisher    = {{ACM}},
  year         = {2024},
  url          = {https://doi.org/10.1145/3626772.3657775},
  doi          = {10.1145/3626772.3657775},
  bibsource    = {dblp computer science bibliography, https://dblp.org}
}

@inproceedings{GFT,
  author       = {Zehong Wang and
                  Zheyuan Zhang and
                  Nitesh V. Chawla and
                  Chuxu Zhang and
                  Yanfang Ye},
  editor       = {Amir Globersons and
                  Lester Mackey and
                  Danielle Belgrave and
                  Angela Fan and
                  Ulrich Paquet and
                  Jakub M. Tomczak and
                  Cheng Zhang},
  title        = {{GFT:} Graph Foundation Model with Transferable Tree Vocabulary},
  booktitle    = {Advances in Neural Information Processing Systems 38: Annual Conference
                  on Neural Information Processing Systems 2024, NeurIPS 2024, Vancouver,
                  BC, Canada, December 10 - 15, 2024},
  year         = {2024},
  url          = {http://papers.nips.cc/paper\_files/paper/2024/hash/c23ccf9eedf87e4380e92b75b24955bb-Abstract-Conference.html},
  bibsource    = {dblp computer science bibliography, https://dblp.org}
}

@inproceedings{LLM-GNN,
  author       = {Zhikai Chen and
                  Haitao Mao and
                  Hongzhi Wen and
                  Haoyu Han and
                  Wei Jin and
                  Haiyang Zhang and
                  Hui Liu and
                  Jiliang Tang},
  title        = {Label-free Node Classification on Graphs with Large Language Models
                  (LLMs)},
  booktitle    = {The Twelfth International Conference on Learning Representations,
                  {ICLR} 2024, Vienna, Austria, May 7-11, 2024},
  publisher    = {OpenReview.net},
  year         = {2024},
  url          = {https://openreview.net/forum?id=hESD2NJFg8},
  bibsource    = {dblp computer science bibliography, https://dblp.org}
}

@inproceedings{LLM-BP,
title={Generalization Principles for Inference over Text-Attributed Graphs  with Large Language Models},
author={Haoyu Peter Wang and Shikun Liu and Rongzhe Wei and Pan Li},
booktitle={Forty-second International Conference on Machine Learning},
year={2025},
url={https://openreview.net/forum?id=dfOqiHuklY}
}

@article{MMGPL,
  title={Mmgpl: Multimodal medical data analysis with graph prompt learning},
  author={Peng, Liang and Cai, Songyue and Wu, Zongqian and Shang, Huifang and Zhu, Xiaofeng and Li, Xiaoxiao},
  journal={Medical Image Analysis},
  volume={97},
  pages={103225},
  year={2024},
  publisher={Elsevier}
}

@inproceedings{GFMSurvey_Position,
  author       = {Haitao Mao and
                  Zhikai Chen and
                  Wenzhuo Tang and
                  Jianan Zhao and
                  Yao Ma and
                  Tong Zhao and
                  Neil Shah and
                  Mikhail Galkin and
                  Jiliang Tang},
  title        = {Position: Graph Foundation Models Are Already Here},
  booktitle    = {Forty-first International Conference on Machine Learning, {ICML} 2024,
                  Vienna, Austria, July 21-27, 2024},
  publisher    = {OpenReview.net},
  year         = {2024},
  url          = {https://openreview.net/forum?id=Edz0QXKKAo},
  bibsource    = {dblp computer science bibliography, https://dblp.org}
}

@article{GraphSSLSurvey,
  author       = {Yixin Liu and
                  Ming Jin and
                  Shirui Pan and
                  Chuan Zhou and
                  Yu Zheng and
                  Feng Xia and
                  Philip S. Yu},
  title        = {Graph Self-Supervised Learning: {A} Survey},
  journal      = {{IEEE} Trans. Knowl. Data Eng.},
  volume       = {35},
  number       = {6},
  pages        = {5879--5900},
  year         = {2023},
  url          = {https://doi.org/10.1109/TKDE.2022.3172903},
  doi          = {10.1109/TKDE.2022.3172903},
  bibsource    = {dblp computer science bibliography, https://dblp.org}
}

@article{GraphAgent,
  title={GraphAgent: Exploiting Large Language Models for Interpretable Learning on Text-attributed Graphs},
  author={Yu, Xinmiao and Qu, Meng and Feng, Xiaocheng and Qin, Bing},
  journal={OpenReview preprint},
  year={2024},
  url={https://openreview.net/forum?id=L3jATpVEGv}
}

@article{DBLP:journals/ijon/WangHQFX20,
  author       = {Jinguang Wang and
                  Jun Hu and
                  Shengsheng Qian and
                  Quan Fang and
                  Changsheng Xu},
  title        = {Multimodal graph convolutional networks for high quality content recognition},
  journal      = {Neurocomputing},
  volume       = {412},
  pages        = {42--51},
  year         = {2020},
  url          = {https://doi.org/10.1016/j.neucom.2020.04.145},
  doi          = {10.1016/J.NEUCOM.2020.04.145},
  bibsource    = {dblp computer science bibliography, https://dblp.org}
}

@article{DBLP:journals/ipm/TaoWWHHC20,
  author       = {Zhulin Tao and
                  Yinwei Wei and
                  Xiang Wang and
                  Xiangnan He and
                  Xianglin Huang and
                  Tat{-}Seng Chua},
  title        = {{MGAT:} Multimodal Graph Attention Network for Recommendation},
  journal      = {Inf. Process. Manag.},
  volume       = {57},
  number       = {5},
  pages        = {102277},
  year         = {2020},
  url          = {https://doi.org/10.1016/j.ipm.2020.102277},
  doi          = {10.1016/J.IPM.2020.102277},
  bibsource    = {dblp computer science bibliography, https://dblp.org}
}

@inproceedings{DBLP:conf/icml/NingFWXH25,
  author       = {Xuying Ning and
                  Dongqi Fu and
                  Tianxin Wei and
                  Wujiang Xu and
                  Jingrui He},
  editor       = {Aarti Singh and
                  Maryam Fazel and
                  Daniel Hsu and
                  Simon Lacoste{-}Julien and
                  Felix Berkenkamp and
                  Tegan Maharaj and
                  Kiri Wagstaff and
                  Jerry Zhu},
  title        = {Graph4MM: Weaving Multimodal Learning with Structural Information},
  booktitle    = {Forty-second International Conference on Machine Learning, {ICML}
                  2025, Vancouver, BC, Canada, July 13-19, 2025},
  series       = {Proceedings of Machine Learning Research},
  volume       = {267},
  publisher    = {{PMLR} / OpenReview.net},
  year         = {2025},
  url          = {https://proceedings.mlr.press/v267/ning25a.html},
  bibsource    = {dblp computer science bibliography, https://dblp.org}
}

@article{DBLP:journals/corr/abs-2506-02568,
  author       = {Dongzhe Fan and
                  Yi Fang and
                  Jiajin Liu and
                  Djellel Difallah and
                  Qiaoyu Tan},
  title        = {MLaGA: Multimodal Large Language and Graph Assistant},
  journal      = {CoRR},
  volume       = {abs/2506.02568},
  year         = {2025},
  url          = {https://doi.org/10.48550/arXiv.2506.02568},
  doi          = {10.48550/ARXIV.2506.02568},
  eprinttype   = {arXiv},
  eprint       = {2506.02568},
  bibsource    = {dblp computer science bibliography, https://dblp.org}
}

@article{DBLP:journals/corr/abs-2603-05181,
  author       = {Yuanfu Sun and
                  Kang Li and
                  Pengkang Guo and
                  Jiajin Liu and
                  Qiaoyu Tan},
  title        = {Mario: Multimodal Graph Reasoning with Large Language Models},
  journal      = {CoRR},
  volume       = {abs/2603.05181},
  year         = {2026},
  url          = {https://doi.org/10.48550/arXiv.2603.05181},
  doi          = {10.48550/ARXIV.2603.05181},
  eprinttype   = {arXiv},
  eprint       = {2603.05181},
  bibsource    = {dblp computer science bibliography, https://dblp.org}
}

@article{ektefaie2023multimodal,
  title={Multimodal learning with graphs},
  author={Ektefaie, Yasha and Dasoulas, George and Noori, Ayush and Farhat, Maha and Zitnik, Marinka},
  journal={Nature Machine Intelligence},
  volume={5},
  number={4},
  pages={340--350},
  year={2023},
  publisher={Nature Publishing Group UK London}
}

@inproceedings{mcqueen1967some,
  title={Some methods of classification and analysis of multivariate observations},
  author={McQueen, James B},
  booktitle={Proc. of 5th Berkeley Symposium on Math. Stat. and Prob.},
  pages={281--297},
  year={1967}
}

@article{sparck1972statistical,
  title={A statistical interpretation of term specificity and its application in retrieval},
  author={Sparck Jones, Karen},
  journal={Journal of documentation},
  volume={28},
  number={1},
  pages={11--21},
  year={1972},
  publisher={MCB UP Ltd}
}

@article{van2008visualizing,
  title={Visualizing data using t-SNE.},
  author={Van der Maaten, Laurens and Hinton, Geoffrey},
  journal={Journal of machine learning research},
  volume={9},
  number={11},
  year={2008}
}

@article{pearson1901liii,
  title={LIII. On lines and planes of closest fit to systems of points in space},
  author={Pearson, Karl},
  journal={The London, Edinburgh, and Dublin philosophical magazine and journal of science},
  volume={2},
  number={11},
  pages={559--572},
  year={1901},
  publisher={Taylor \& Francis}
}

@inproceedings{DBLP:conf/kdd/0004L0YHLZ0W25,
  author       = {Hao Yan and
                  Chaozhuo Li and
                  Jun Yin and
                  Zhigang Yu and
                  Weihao Han and
                  Mingzheng Li and
                  Zhengxin Zeng and
                  Hao Sun and
                  Senzhang Wang},
  editor       = {Luiza Antonie and
                  Jian Pei and
                  Xiaohui Yu and
                  Flavio Chierichetti and
                  Hady W. Lauw and
                  Yizhou Sun and
                  Srinivasan Parthasarathy},
  title        = {When Graph Meets Multimodal: Benchmarking and Meditating on Multimodal
                  Attributed Graph Learning},
  booktitle    = {Proceedings of the 31st {ACM} {SIGKDD} Conference on Knowledge Discovery
                  and Data Mining, V.2, {KDD} 2025, Toronto ON, Canada, August 3-7,
                  2025},
  pages        = {5842--5853},
  publisher    = {{ACM}},
  year         = {2025},
  url          = {https://doi.org/10.1145/3711896.3737404},
  doi          = {10.1145/3711896.3737404},
  bibsource    = {dblp computer science bibliography, https://dblp.org}
}

\clearpage
\appendix

\section{Algorithm}\label{algorithm}
\begin{algorithm}[!t]
\caption{\textsc{CHARM} Training and Zero-Shot Inference}
\label{alg:charm}
\small
\begin{algorithmic}[1]

\State \textbf{Input:} Multimodal graphs
$\{\mathcal{G}_d=(\mathcal{V}_d,\mathcal{E}_d)\}_{d\in\mathcal D}$;
\State \hspace{1em} source-domain NC and LP samples
$\mathcal{D}_{\mathrm{src}}^{\mathrm{NC}}$ and
$\mathcal{D}_{\mathrm{src}}^{\mathrm{LP}}$, frozen encoders and LLM.
\State \textbf{Output:} Parameters $\theta$ and prediction
$\widehat{y}$.

\Statex
\State \textbf{(1) Synthetic hierarchical graph construction}

\For{each domain $d\in\mathcal D$}
    \State Compute $\mathbf{x}_v^{t}\leftarrow
    \mathrm{Enc}_{t}(s_v^{t})$ and
    $\mathbf{x}_v^{i}\leftarrow
    \mathrm{Enc}_{i}(s_v^{i})$ for all $v\in\mathcal V_d$.

    \State $\mathcal{A}_d^{(1)}
    \leftarrow
    \mathrm{KMeans}
    (\{\mathbf{x}_v^{t}:v\in\mathcal V_d\})$

    \State $\mathcal{A}_d^{(2)}
    \leftarrow
    \mathrm{KMeans}
    (\{\mathbf{a}^{(1),t}:a^{(1)}\in\mathcal A_d^{(1)}\})$
\EndFor

\State $\mathcal{P}\leftarrow
\mathrm{KMeans}
\big(\bigcup_{d\in\mathcal D}
\{\mathbf{a}^{(2),t}:a^{(2)}\in\mathcal A_d^{(2)}\}\big)$

\For{each synthetic cluster $C$}
    \State $\mathbf{a}_C^{t}\leftarrow\mathrm{Enc}_{t}(s_C)$

    \State $\displaystyle
    \alpha_v\leftarrow
    \frac{\exp(\cos(\mathbf{a}_C^{t},\mathbf{x}_v^{t})/\tau)}
    {\sum_{u\in C}
    \exp(\cos(\mathbf{a}_C^{t},\mathbf{x}_u^{t})/\tau)}$

    \State $\mathbf{a}_C^{i}\leftarrow
    \sum_{v\in C}\alpha_v\mathbf{x}_v^{i}$
\EndFor

\Statex
\State \textbf{(2) Modality-complementary bridge construction}

\For{each raw node $v_i$}
    \State Obtain PPR candidates
    $\mathcal{N}_{\mathrm{PPR}}(v_i)$.

    \For{each $v_j\in\mathcal{N}_{\mathrm{PPR}}(v_i)$}
        \State $s_t(v_i,v_j)\leftarrow
        \cos(\mathbf{x}_{v_i}^{t},\mathbf{x}_{v_j}^{t})$,
        $\quad
        s_i(v_i,v_j)\leftarrow
        \cos(\mathbf{x}_{v_i}^{i},\mathbf{x}_{v_j}^{i})$

        \State $\displaystyle
        r_{ij}\leftarrow
        \frac{
        |\mathcal{N}_{\mathrm{PPR}}(v_i)
        \cap\mathcal{N}_{\mathrm{PPR}}(v_j)|
        }{
        \min(
        |\mathcal{N}_{\mathrm{PPR}}(v_i)|,
        |\mathcal{N}_{\mathrm{PPR}}(v_j)|)
        }$
    \EndFor

    \State Retain at most $K_b$ candidates satisfying
    \Statex \hspace{\algorithmicindent}
    $\big(s_t\geq\tau_t^h,\ s_i\leq\tau_i^l\big)$
    or
    $\big(s_i\geq\tau_i^h,\ s_t\leq\tau_t^l\big)$,
    together with the minimum PPR-overlap criterion, ranked by $r_{ij}$.
\EndFor

\Statex
\State \textbf{(3) Joint graph context encoding and source training}

\State $\mathcal{D}_{\mathrm{src}}^{\mathrm{joint}}\leftarrow
\mathcal{D}_{\mathrm{src}}^{\mathrm{NC}}\cup
\mathcal{D}_{\mathrm{src}}^{\mathrm{LP}}$

\While{training schedule not complete}
    \For{each task-tagged $(t,x,y)\in\mathcal{D}_{\mathrm{src}}^{\mathrm{joint}}$}
        \State Construct the context of each center node $v_c$:
        \Statex \hspace{\algorithmicindent}
        $\displaystyle
        \mathcal{C}(v_c)\leftarrow
        \{v_c\}\cup\mathcal{N}_{\mathrm{PPR}}(v_c)
        \cup\mathcal{A}(v_c)\cup\mathcal{G}(v_c)
        \cup\mathcal{B}(v_c)$

        \For{each $u\in\mathcal{C}(v_c)$}
            \State $\mathbf{h}_u^{t}\leftarrow
            \mathbf{W}_t\mathbf{x}_u^{t}$,
            $\quad
            \mathbf{h}_u^{i}\leftarrow
            \mathbf{W}_i\mathbf{x}_u^{i}$

            \State $\mathbf{q}_u\leftarrow
            [\mathbf{h}_u^{t};
            \mathbf{h}_u^{i};
            \mathbf{h}_u^{t}\odot\mathbf{h}_u^{i};
            |\mathbf{h}_u^{t}-\mathbf{h}_u^{i}|]$

            \State $g_u\leftarrow
            \sigma(\mathrm{MLP}_{g}(\mathbf{q}_u))$

            \State $\mathbf{f}_u\leftarrow
            \mathrm{MLP}_{f}
            ([g_u\mathbf{h}_u^{t};
            (1-g_u)\mathbf{h}_u^{i}])$
        \EndFor

        \State $\widetilde{\mathbf{A}}_c\leftarrow
        (\mathbf{A}_c^{\mathrm{ori}}
        \vee\mathbf{A}_c^{\mathrm{hier}}
        \vee\mathbf{A}_c^{\mathrm{br}})+\mathbf{I}$

        \State $\mathbf{H}^{(\ell+1)}\leftarrow
        \sigma(
        \widehat{\mathbf{A}}_c
        \mathbf{H}^{(\ell)}
        \mathbf{W}^{(\ell)})$

        \State $\mathbf{h}_c^{\mathrm{out}}\leftarrow
        (1-\gamma)\mathbf{H}_c^{(0)}
        +\gamma\mathbf{H}_c^{(L)}$

        \State Obtain graph tokens
        $\mathbf{S}_{x}$ after layer normalization.

        \State $\mathcal{L}\leftarrow
        -\log P_{\phi}
        (y\mid\mathbf{S}_{x}(\theta),\mathbf{T}_{x}^{t})$

        \State Update $\theta$ by minimizing $\mathcal{L}$.
    \EndFor
\EndWhile

\Statex
\State \textbf{(4) Target-domain zero-shot inference}

\State Construct $\mathbf{S}_{x}$ for each target query without
updating $\theta$.

\State $\widehat{y}\leftarrow
\operatorname*{arg\,max}_{y}
P_{\phi}(y\mid\mathbf{S}_{x}(\theta),\mathbf{T}_{x})$

\State \textbf{return} $\theta$ and $\widehat{y}$

\end{algorithmic}
\end{algorithm}

Algorithm~\ref{alg:charm} summarizes the training and zero-shot
inference procedure of \textsc{CHARM}. The hierarchy, PPR neighborhoods, and
bridge relations are constructed once as label-free graph preprocessing,
whereas the compact context and graph-token sequence are generated for each
query.

\paragraph{Step (1): hierarchical graph construction.}
Frozen encoders produce normalized text and image features for every raw node.
Within each domain, text features are clustered into Layer-1 and Layer-2
anchors, and the Layer-2 anchors from the three domains are further clustered
into shared global anchors. Each raw node is therefore connected through
$v\rightarrow a^{(1)}\rightarrow a^{(2)}\rightarrow p$. For a synthetic
cluster $C$, its label-free summary $s_C$ provides the textual feature
$\mathbf{a}_C^t$, while text-guided aggregation produces the visual feature
$\mathbf{a}_C^i$. No category label is involved in this construction.

\paragraph{Step (2): bridge construction.}
For each raw node $v_i$, bridge candidates are restricted to
$\mathcal{N}_{\mathrm{PPR}}(v_i)$. A candidate pair is retained when text and
image similarities exhibit a clear reliability asymmetry and its
PPR-neighborhood consistency $r_{ij}$ is sufficiently high. Each node keeps at
most $K_b$ candidates ranked by $r_{ij}$, yielding bridge relations that preserve useful evidence
visible in only one modality while filtering structurally weak matches.

\paragraph{Step (3): joint context encoding and source training.}
The NC and LP examples from the source domains are combined into a single
joint training set. For each task-tagged query, \textsc{CHARM} retrieves the
center node, PPR neighbors, Layer-1 anchors, Layer-2 anchors, global anchors,
and bridge neighbors to form $\mathcal{C}(v_c)$. The reliability gate
adaptively fuses the projected text and image features of every context item.
The fused states are propagated over the union of original, hierarchical, and
bridge relations, after which the center-protection gate combines the initial
and propagated center states. The resulting sequence $\mathbf{S}_x$ is
inserted into the frozen LLM with the corresponding task prompt, and the shared
graph context encoder and projection modules are jointly optimized over both
tasks.

\paragraph{Step (4): zero-shot inference.}
The trained graph context encoder is directly applied to the target graph
without target-domain parameter updates. The frozen LLM scores the target
category options for node classification or the \emph{Yes}/\emph{No} options
for link prediction, and returns the answer with the highest conditional
probability.

\section{Experimental Details}
\label{app:protocol}

\subsection{Zero-Shot Evaluation Protocol}
\label{app:zero_shot_protocol}

We consider two three-domain groups, Grocery--Movies--Toys (GMT) and Arts--Beauty--CD (ABC). Each reported transfer uses two complete graphs as source domains and one graph as the target domain. For example, Grocery+Movies$\rightarrow$Toys indicates that model parameters are optimized using Grocery and Movies and then directly evaluated on Toys. For the GMT benchmark group, Grocery, Movies, and Toys are each evaluated once as the unseen target domain. For the ABC group, Arts and CD are evaluated as unseen target domains, while Beauty is used only as a source domain.

The target graph is available before prediction only through its unlabeled text, images, and topology. These inputs are used to build the semantic hierarchy, cache PPR neighborhoods, discover bridge relations, and retrieve query-specific contexts. Target labels are never used for clustering, anchor assignment, global-anchor construction or bridge discovery. Target category names are exposed only as candidate answers for node classification, which is
necessary to define the open-vocabulary output space. This protocol is therefore zero-shot transfer: the unlabeled target graph is
visible, but no target supervision is available.

For all methods, source and target partitions are fixed before training and shared across comparisons. Validation data are used only on the source side. For \textsc{CHARM}, each source configuration produces one checkpoint jointly trained on its NC and LP examples, and this same checkpoint is applied unchanged to both tasks in the held-out target domain. The hierarchy and bridge graph are also reconstructed from the target attributes without referring to target performance.

\subsection{Datasets and Data Construction}
\label{app:datasets}

Table~\ref{tab:app_dataset_stats} reports the graph sizes used in the experiments. Grocery, Movies, and Toys are adopted from the public multimodal Amazon product-graph benchmark~\cite{DBLP:conf/kdd/0004L0YHLZ0W25}. In these graphs, a node represents an Amazon item, the text modality contains its title and available description, the visual modality contains the associated product image, and an edge denotes an item--item relation derived from co-purchase behavior. The prediction label is
the product category. We use the released graph structure and multimodal attributes without resampling these three domains.

Arts, Beauty, and CD are category-stratified induced subgraphs constructed from the original Amazon product data. We first retain records with valid node identifiers, category information, textual metadata, and associated images. Nodes are then sampled independently within each category so that the retained
subsets approximately preserve the original class proportions rather than forming artificially balanced graphs. This design keeps the naturally long-tailed category distribution. For example, among the 13 Arts classes, large categories such as Sewing, Crafting, and Painting contain approximately 1,500--2,400 sampled nodes, whereas tail categories such as Sports, Collectible Coins, and Org contain approximately 150--200 nodes. The final
Arts graph contains 11,640 nodes sampled from 775,900 source rows. 

After node sampling, each graph is formed as an induced co-purchase subgraph. An original edge is retained only when both endpoints belong to the sampled node set, after which identifiers are remapped to a contiguous range. Therefore, all reported edges connect retained nodes; no full-graph node is implicitly included. The comparatively large edge counts of Beauty and CD
reflect denser co-purchase relations among the sampled products. Within each class, nodes are assigned to training, validation, and test partitions with an approximately $60\%/20\%/20\%$ stratified ratio, ensuring that both head and tail classes remain represented across partitions.

\begin{table}[t]
\centering
\caption{Statistics of the six multimodal graphs.}
\label{tab:app_dataset_stats}
\small
\setlength{\tabcolsep}{4.0pt}
\renewcommand{\arraystretch}{1.04}
\begin{tabular}{lrrr}
\toprule
Dataset & \#Nodes & \#Edges & \#Classes \\
\midrule
Grocery & 17,074 & 95,250 & 20 \\
Movies  & 16,672 & 137,067 & 20 \\
Toys    & 20,695 & 66,680 & 18 \\
Arts    & 11,640 & 884,312 & 13 \\
Beauty  & 11,874 & 3,779,088 & 9 \\
CD      & 11,851 & 7,168,792 & 19 \\
\bottomrule
\end{tabular}
\end{table}

\subsection{Tasks, Splits, and Metrics}
\label{app:tasks_metrics}

\paragraph{Node classification.}
Each target instance contains one center node and the complete set of category
names defined by the target dataset. The graph context encoder constructs the
target-dependent token sequence, and the frozen LLM evaluates every category
as a candidate answer. No classifier is fitted to target embeddings and no labeled target example is inserted into the prompt. We report Accuracy and Macro-F1. Accuracy measures the proportion of correctly classified nodes,
whereas Macro-F1 assigns equal weight to every class and is therefore important for the long-tailed category distributions of the product graphs. A large gap between the two metrics can indicate that performance is concentrated in frequent head classes while tail-class coverage remains limited.

\paragraph{Link prediction.}
Positive examples are observed co-purchase edges and negative examples are randomly sampled node pairs without an edge. Each graph provides 10,000 source-domain training pairs and 3,200 target-domain test pairs, with equal numbers of positive and negative instances. The source training, source validation, and target test pairs are disjoint. For a positive query
$(v_i,v_j)$, the queried edge is removed before PPR computation, bridge filtering, context retrieval, and context-adjacency construction. The model therefore cannot recover the answer by directly observing the evaluated edge. The frozen LLM chooses between \emph{Yes} and \emph{No}, and we report Accuracy and F1.

\subsection{Implementation Details}
\label{app:implementation}

\paragraph{Running environment.}
All experiments are conducted on a Linux server with 125 GiB RAM and 8 NVIDIA GeForce
RTX 5090 GPUs (32 GB memory each). We use Python 3.10.20, PyTorch 2.11.0 (CUDA 12.8), and Transformers 4.40.0. LLM-based models are loaded in float16
precision. A single GPU is used for each reported training run. Gradient checkpointing is enabled when necessary to remain within
the 32\,GB memory budget. Cached multimodal features, hierarchy assignments, PPR candidates, and bridge relations are reused across epochs so that online training does not repeatedly execute the offline preprocessing pipeline.

\paragraph{Baseline implementations.}
All open-source baselines are run using their publicly released repositories. We follow the optimizer, number of training epochs, neighborhood construction, prompt format, and other settings reported in the
corresponding papers or released configurations. Thus, the reported final training times reflect each method's original schedule rather than forcing all models to use an identical epoch count. The same splits and target
answer spaces are used for all methods. For traditional GNNs, self-supervised graph methods, and GNN-based foundation models that require a
single node feature matrix, frozen CLIP text and image features are averaged as
$(\mathbf{x}^{t}_v+\mathbf{x}^{i}_v)/2$. LLM-based baselines follow their released graph-input and inference protocols, while the target domain remains free of label supervision.

\paragraph{Prompt construction.}
Each \texttt{<graph>} placeholder is replaced at runtime by the graph soft-token
sequence produced for the corresponding center node. The textual prompt identifies
the roles of the retrieved evidence, whereas the continuous graph tokens retain their
multimodal and structural representations. No labeled target-domain node or target-domain
demonstration is included. We use separate prompt templates for node classification
and link prediction, and detail the main NC fields below.

\paragraph{Node classification prompt.}
The NC prompt contains one \texttt{<graph>} placeholder for the queried node and its retrieved context. To keep the interface concise, the textual part provides only the center description, hierarchical summaries, modality-complementary bridge-neighbor descriptions, and the target-domain candidate labels. The continuous graph tokens retain the corresponding multimodal features and graph relations. The main template is:

\begin{lstlisting}[style=promptstyle]
### Human:
[TASK] Classify the center node in the following multimodal graph context.

[GRAPH CONTEXT]
<graph>

[CENTER NODE]
Text={center_text}

[HIERARCHICAL SEMANTIC CONTEXT]
Layer-1 summary={L1_summary}
Layer-2 summary={L2_summary}
Global anchor summaries={G_summaries}
Modality-complementary bridge neighbors={bridge_neighbors}

[CANDIDATE LABELS]
({option_1}) {label_1}: {label_description_1}
({option_2}) {label_2}: {label_description_2}
[repeat for all target-domain labels]

[OUTPUT FORMAT]
Return exactly one option letter.

### Assistant:
{ground_truth_option}
\end{lstlisting}
During training, \texttt{ground\_truth\_option} is the supervised assistant target; at inference, the response is left empty and the fixed option letters are scored. The main textual fields are summarized in Table~\ref{tab:app_nc_prompt_fields}.

\begin{table*}[t]
\centering
\caption{Main textual fields in the NC prompt.}
\label{tab:app_nc_prompt_fields}
\footnotesize
\setlength{\tabcolsep}{4pt}
\renewcommand{\arraystretch}{1.04}
\begin{tabularx}{\textwidth}{@{}p{0.22\textwidth}X@{}}
\toprule
Field & Role in the prompt \\
\midrule
\texttt{center\_text} & The title and description of the queried product. It is the primary instance-level textual evidence. \\
\texttt{L1\_summary} & A label-free summary of the assigned Layer-1 anchor, representing a relatively fine-grained semantic group. \\
\texttt{L2\_summary} & A label-free summary of the corresponding Layer-2 anchor, providing a broader in-domain semantic topic. \\
\texttt{G\_summaries} & Summaries of the linked global anchors, providing higher-level semantic references shared across domains. \\
\texttt{bridge\_neighbors} & Short descriptions of structurally supported neighbors that are strongly related in one modality but weakly related in the other. \\
\texttt{label\_name} and \texttt{label\_description} & The target-domain candidate category and its semantic description. Each candidate is assigned a fixed option letter. \\
\bottomrule
\end{tabularx}
\end{table*}

\paragraph{Link prediction prompt.}
The LP prompt uses two \texttt{<graph>} placeholders, one for each queried product. Each product reuses the center, hierarchical, and modality-complementary bridge-neighbor fields defined for NC. The simplified template is:

\begin{lstlisting}[style=promptstyle]
### Human:
[TASK] Decide whether Product A and Product B are linked in the
{domain_name} graph.

[GRAPH CONTEXTS]
Target Product A=<graph>
Target Product B=<graph>

[TARGET PAIR]
Product A text={product_A_text}
Product B text={product_B_text}

[HIERARCHICAL SEMANTIC CONTEXT]
Product A: Layer-1={A_L1_summary};
Layer-2={A_L2_summary}; Global={A_G_summaries}
Modality-complementary bridge neighbors={bridge_neighbors}
Product B: Layer-1={B_L1_summary};
Layer-2={B_L2_summary}; Global={B_G_summaries}
Modality-complementary bridge neighbors={bridge_neighbors}

[OUTPUT FORMAT]
Return exactly Yes or No.

### Assistant:
{answer}
\end{lstlisting}
During training, \texttt{answer} is the supervised \emph{Yes}/\emph{No} target; at inference, the two answer logits are compared directly.

\paragraph{Frozen feature extraction and synthetic hierarchy.}
We use the frozen \texttt{openai/clip-vit-large-patch14-336} encoder and take its penultimate-layer features, producing 1,024-dimensional text and image representations that are $L_2$-normalized and cached before training. Synthetic nodes use the same two-channel format. For each Layer-1 cluster, up to 80 member titles and descriptions are processed by TF-IDF with 1--3-grams and stop-word removal; the top eight terms form the template content, and the first two form the semantic label. Layer-2 summaries apply TF-IDF to child Layer-1 summaries using the top ten terms and combine the most frequent child labels. Global anchors follow the Layer-2 rule with a cross-dataset template. No LLM or category label is used to generate any summary. Within each domain, the expected Layer-1 cluster size is 64 raw nodes. Approximately eight Layer-1 anchors are grouped into one Layer-2 anchor, and Layer-2 anchors from the three domains in a benchmark group are jointly clustered into 24 global anchors. Each Layer-2 anchor is connected to its three nearest global anchors, including its assigned global anchor. The hierarchy is constructed once per benchmark group from the available unlabeled graph attributes using the same fixed settings.

\paragraph{Graph context construction.}
For every raw node, we cache its top-128 PPR candidates using teleport probability $\alpha=0.15$, tolerance $10^{-4}$, and at most 100,000 push steps. Textual and visual similarities are evaluated only within this candidate set. For each dataset and modality, the high and low thresholds are the 75th and 35th percentiles estimated from 200,000 randomly sampled pairs. The PPR-neighborhood overlap $r_{ij}$ is then used to remove structurally inconsistent matches, and each node retains at most four valid bridge neighbors ranked by $r_{ij}$.

At query time, the local component contains at most eight raw neighbors with the highest PPR scores. Hierarchical retrieval follows the precomputed assignment path: the center node's assigned Layer-1 anchor, its corresponding Layer-2 anchor, and up to three global anchors linked to that Layer-2 anchor are retained. The bridge component contains at most four modality-complementary neighbors obtained through the filtering procedure described above.

The final context consists of the center node, at most eight PPR
neighbors, one selected Layer-1 anchor, one selected Layer-2 anchor, at
most three global anchors, and at most four bridge neighbors.
Duplicate items are removed while their roles are preserved, yielding
at most 18 context items. The retained items are deterministically
ordered as the center, PPR neighbors, bridge neighbors, and
hierarchical evidence. Original, hierarchical, and bridge relations
among these items define the compact adjacency used by the graph
context encoder.
\paragraph{Graph context encoder and optimization.}
The text and image channels are projected separately from 1,024 dimensions to the 4,096-dimensional hidden space of the frozen \texttt{lmsys/vicuna-7b-v1.5} backbone. A shared node-wise reliability gate is evaluated independently for every raw or synthetic context item, allowing the relative modality contribution to vary across nodes and domains. The fused states are processed by two compact-graph propagation layers. The center-protection gate then combines the initial and propagated states of the query center, while non-center items use their final propagated states. Layer normalization produces the graph soft tokens inserted at the graph placeholder of the LLM input. Trainable parameters consist of the modality projectors, reliability gate, fusion MLP, propagation layers, and center-protection parameters. NC and LP examples are optimized together in a single training run with shared parameters and a common optimizer. The tasks differ only in their prompt templates and answer spaces, and the same resulting checkpoint is used for all reported NC and LP evaluations within a transfer setting.

\paragraph{Training objective and option scoring.}
Source-domain training jointly minimizes the standard next-token prediction loss over NC and LP examples. For NC, the target is the correct option letter; for LP, it is \emph{Yes} or \emph{No}. Both losses update the same graph context encoder and projection modules within a single training schedule. At inference, we use an option-logit formulation rather than sequence likelihood. Let $a_k$ be the fixed letter assigned to candidate $k$, and let $\mathcal{I}(a_k)$ contain all token IDs representing that letter. From the final-position logits $\boldsymbol{\ell}(x)$, we compute
\begin{equation}
s_k(x)=\operatorname{logsumexp}_{z\in\mathcal{I}(a_k)}\ell_z(x).
\end{equation}
The candidate with the largest score is returned. Candidate ordering is fixed; labels are not shuffled, and no length normalization or target-domain demonstration is applied.

\paragraph{Training time and memory.}
For each method, we report the actual elapsed training time for the Grocery+Movies$\rightarrow$Toys setting, following its original implementation and training schedule. For \textsc{CHARM}, this time covers the single joint NC--LP training run. The training times are 87 minutes for LLaGA, 194 minutes for GraphGPT, 141 minutes for Graph4MM, 279 minutes for TEA-GLM, 349 minutes for MLaGA, and 323 minutes for \textsc{CHARM}. Their corresponding peak GPU-memory usages are 22.74, 22.75, 29.19, 15.83, 17.24, and 28.57\,GB, respectively.

LLaGA achieves the shortest runtime because it reorganizes graph neighborhoods into structure-aware node sequences and mainly learns a projector that maps these sequences into the LLM embedding space. GraphGPT has a similar peak memory but requires more time because it performs
text--graph grounding followed by two-stage graph instruction tuning. TEA-GLM first aligns pretrained GNN representations with the LLM token
embedding space and then learns a linear graph-token projector. Its fixed-number graph tokens and absence of explicit visual processing result
in the lowest peak memory, whereas the additional GNN pretraining and
alignment stage increase the total runtime. Graph4MM applies Hop-Diffused Attention to inject multi-hop structure into textual and visual representations and employs MM-QFormer for cross-modal fusion. Retaining modality-specific states, diffused attention features,
and query tokens simultaneously leads to the highest peak memory, although its shallow fusion module keeps the runtime moderate. MLaGA trains a structure-aware multimodal encoder through joint graph pretraining and subsequently performs multimodal graph instruction tuning.
These two optimization stages lead to the longest runtime. \textsc{CHARM} precomputes and caches the multimodal features, hierarchical
anchors, PPR candidates, and bridge relations, and its online graph propagation is restricted to at most 18 context items. These choices make it faster than GraphGPT, Graph4MM, TEA-GLM, and MLaGA. Its peak memory remains relatively high because the text and image features of every context item are separately projected into the 4,096-dimensional LLM space, followed by
reliability-aware fusion, graph propagation, and graph-token processing. 

\begin{table}[t]
\centering
\caption{Notation used in the main formulation of \textsc{CHARM}.}
\label{tab:app_notation}
\footnotesize
\setlength{\tabcolsep}{2.5pt}
\renewcommand{\arraystretch}{1.03}
\begin{tabularx}{\columnwidth}{@{}p{0.33\columnwidth}X@{}}
\toprule
Symbol & Description \\
\midrule
$\mathcal{D},d$ & Set of graph domains and a domain index. \\
$\mathcal{G}_d=(\mathcal{V}_d,\mathcal{E}_d)$ & Multimodal graph, node set, and edge set of domain $d$. \\
$\mathcal{G}_{\mathrm{syn}}$ & Synthetic hierarchical graph. \\
$\mathbf{x}^{t}_v,\mathbf{x}^{i}_v$ & Frozen text and image features of raw node $v$. \\
$\mathrm{Enc}_t,\mathrm{Enc}_i$ & Frozen text and image encoders. \\
$\boldsymbol{\mu}^{(1)}_{d,k},c(v)$ & Layer-1 centroid $k$ and the assignment of node $v$. \\
$s_1$ & Layer-1 cluster size. \\

$a^{(1)},a^{(2)}$ & Layer-1 anchor, Layer-2 anchor. \\
$p,\mathcal{P}$ & One global anchor and the global-anchor set. \\
$C,s_C$ & A synthetic cluster and its label-free textual summary. \\
$\mathbf{a}^{t}_C,\mathbf{a}^{i}_C$ & Textual and visual features of synthetic cluster $C$. \\
$\alpha_v$ & Visual aggregation weight. \\
$m$ & Number of global anchors linked to a Layer-2 anchor. \\
$s_t(v_i,v_j),s_i(v_i,v_j)$ & Textual and visual cosine similarities of a candidate pair. \\
$\tau_t^h,\tau_t^l,\tau_i^h,\tau_i^l$ & Domain-adaptive high/low thresholds for text and image similarities. \\
$\mathcal{N}_{\mathrm{PPR}}(v)$ & PPR-ranked candidate neighborhood of node $v$. \\
$r_{ij}$ & PPR-neighborhood consistency of nodes $v_i$ and $v_j$. \\
$K_p,K_b$ & Maximum numbers of PPR and bridge neighbors. \\
$v_c,u$ & Center node and an item in its retrieved context. \\
$\mathcal{A}(v_c),\mathcal{G}(v_c),\mathcal{B}(v_c)$ & Retrieved anchors, global anchors, and bridge neighbors. \\
$\mathcal{C}(v_c)$ & Compact context constructed for center node $v_c$. \\
$\mathbf{W}_t,\mathbf{W}_i$ & Modality-specific projection matrices. \\
$\mathbf{h}^{t}_u,\mathbf{h}^{i}_u$ & Projected text and image states of context item $u$. \\
$\mathbf{q}_u$ & Cross-modal interaction descriptor of item $u$. \\
$g_u,\mathrm{MLP}_g$ & Node-wise reliability gate and its gating network. \\
$\mathbf{f}_u,\mathrm{MLP}_f$ & Fused multimodal state and fusion network. \\
$\mathbf{A}^{\mathrm{ori}}_c,\mathbf{A}^{\mathrm{hier}}_c,\mathbf{A}^{\mathrm{br}}_c$ & Context-restricted original, hierarchical, and bridge adjacencies. \\
$\widetilde{\mathbf{A}}_c,\widetilde{\mathbf{D}}_c$ & Context adjacency with self-loops and its degree matrix. \\
$\widehat{\mathbf{A}}_c$ & Symmetrically normalized context adjacency. \\
$\mathbf{H}^{(\ell)},\mathbf{W}^{(\ell)}$ & Context states and propagation weights at layer $\ell$. \\
$L,\sigma(\cdot)$ & Number of propagation layers and activation function. \\
$\gamma,\mathbf{h}^{\mathrm{out}}_c$ & Center-protection coefficient and protected center representation. \\
$\mathbf{s}_n,\mathbf{S}_{v_c}$ & One graph token and the ordered graph-token sequence. \\
$\mathcal{D}_{\mathrm{src}}$ & Source-domain training set. \\
$\mathbf{S}_x,\mathbf{T}_x$ & Graph-token sequence and task input of instance $x$. \\
$P_{\phi},\theta,\mathcal{L}$ & Frozen LLM distribution, trainable graph-module parameters, and training loss. \\
\bottomrule
\end{tabularx}
\end{table}

\section{Visualization Protocols}
\label{app:analysis}

\subsection{Domain Bias and Semantic Relevance Retrieval}
\label{app:motivation_retrieval}

The retrieval  isolates the effect of semantic abstraction from the learned graph context encoder. It asks whether a query from one product domain retrieves units that remain dominated by its own domain or instead discovers
semantically related units from another domain. Only frozen features and the offline hierarchy are used.  For semantic level
$\ell\in\{\mathrm{Raw},L1,L2,G\}$, the candidate pools are
\begin{equation}
\begin{aligned}
\mathcal{U}^{\mathrm{Raw}}&=\mathcal{V}_{G}\cup\mathcal{V}_{M}\cup\mathcal{V}_{T},\\
\mathcal{U}^{L1}&=\mathcal{A}^{L1}_{G}\cup\mathcal{A}^{L1}_{M}\cup\mathcal{A}^{L1}_{T},\\
\mathcal{U}^{L2}&=\mathcal{A}^{L2}_{G}\cup\mathcal{A}^{L2}_{M}\cup\mathcal{A}^{L2}_{T},\\
\mathcal{U}^{G}&=\mathcal{P},
\end{aligned}
\end{equation}
where $G$, $M$, and $T$ denote Grocery, Movies, and Toys. At the raw level, the query representation is the frozen feature of the query node. At Layer-1, Layer-2, and global levels, the query is represented by its assigned semantic unit. After $L_2$ normalization, the top-$K$ retrieved units are
\begin{equation}
\mathcal{N}^{(\ell)}_K(v)=
\operatorname{TopK}_{u\in\mathcal{U}^{(\ell)}\setminus\{q^{(\ell)}(v)\}}
\cos\!\left(\mathbf{z}^{(\ell)}_v,\mathbf{z}^{(\ell)}_u\right).
\end{equation}
The query unit itself is excluded. Anchors at all levels are treated as unique semantic units rather than being replicated for all descendant raw nodes. This avoids artificial duplicate retrievals and similarity ties that
would otherwise overcount large clusters.

Let $\mathcal{M}_u$ denote the raw members represented by semantic unit $u$.
Because a global anchor may contain nodes from several domains, its contribution is weighted by the member-domain distribution
\begin{equation}
\pi_u(d)=\frac{|\mathcal{M}_u\cap\mathcal{V}_d|}{|\mathcal{M}_u|}.
\end{equation}
The same-domain ratio is
\begin{equation}
R_{\mathrm{domain}}^{(\ell)}=
\frac{1}{|\mathcal{Q}|}\sum_{v\in\mathcal{Q}}\frac{1}{K}
\sum_{u\in\mathcal{N}^{(\ell)}_K(v)}\pi_u(d_v).
\end{equation}
For a raw node or a domain-specific anchor, $\pi_u(d)$ reduces to a binary domain indicator. For a shared global anchor, it gives fractional credit according to the global anchor's actual member composition. A lower value indicates that
the retrieved set is less dominated by the query domain and therefore exposes more cross-domain references. The semantic-relevance metric evaluates whether this increased
cross-domain retrieval is meaningful rather than arbitrary. Category labels are encoded only after retrieval and are used solely as evaluation metadata. For a candidate unit $u$, the Grocery-associated category representation is
\begin{equation}
\overline{\mathbf{e}}_{u,G}=
\operatorname{norm}\!\left(
\sum_{w\in\mathcal{M}_u\cap\mathcal{V}_G}\mathbf{e}(y_w)
\right),
\end{equation}
and cross-domain semantic relevance is
\begin{equation}
R_{\mathrm{semantic}}^{(\ell)}=
\frac{1}{|\mathcal{Q}|}\sum_{v\in\mathcal{Q}}\frac{1}{K}
\sum_{u\in\mathcal{N}^{(\ell)}_K(v)}
\mathbf{1}[|\mathcal{M}_u\cap\mathcal{V}_G|>0]
\cos\!\left(\mathbf{e}(y_v),\overline{\mathbf{e}}_{u,G}\right).
\end{equation}
The domain-ratio metric and semantic-relevance metric should be interpreted
together. Reducing domain concentration alone would not be useful if the retrieved units were unrelated; the observed combination of lower domain bias and higher semantic relevance instead indicates that the hierarchy uncovers transferable cross-domain concepts. Labels never affect candidate construction, feature similarity, or retrieval ordering.

\subsection{Additional Visualization Protocols}
\label{app:global_token_geometry}

\paragraph{Text--image similarity distribution.}
For every raw node $v_i$, we pair it with all nodes in its cached top-128 PPR
candidate set and compute
$(s_t(v_i,v_j),s_i(v_i,v_j))$. Restricting the plot to PPR candidates avoids
visualizing arbitrary all-pairs feature coincidences and focuses the analysis
on structurally relevant relations. The same axis range and threshold
interpretation are used across domains. When the number of points is too large
for legible rendering, a fixed-seed uniform subset is used only for plotting;
bridge construction still uses the complete cached candidate set. Points in
the text-strong/image-weak and image-strong/text-weak regions directly
illustrate why an agreement-only retrieval rule would discard useful
one-modality evidence.

\paragraph{t-SNE representation analysis.}
The comparison uses exactly the same sampled target-node indices for frozen
CLIP, MLaGA, and \textsc{CHARM}. Frozen CLIP provides the raw multimodal
baseline, MLaGA provides the node representations released or reproduced under
its official protocol, and \textsc{CHARM} uses the final center graph tokens.
Each representation set is $L_2$-normalized before dimensionality reduction.
t-SNE~\cite{van2008visualizing} is fitted separately for each method because the representation spaces
are not aligned, but the random seed, perplexity, optimization iterations, and
sample indices are held fixed. Class labels are used only to color the final
2D points and do not enter fitting. The visualization therefore compares the
class organization induced by each representation rather than forcing the
methods into a common coordinate system.

\begin{figure}[t]
    \centering
    \includegraphics[width=\columnwidth]{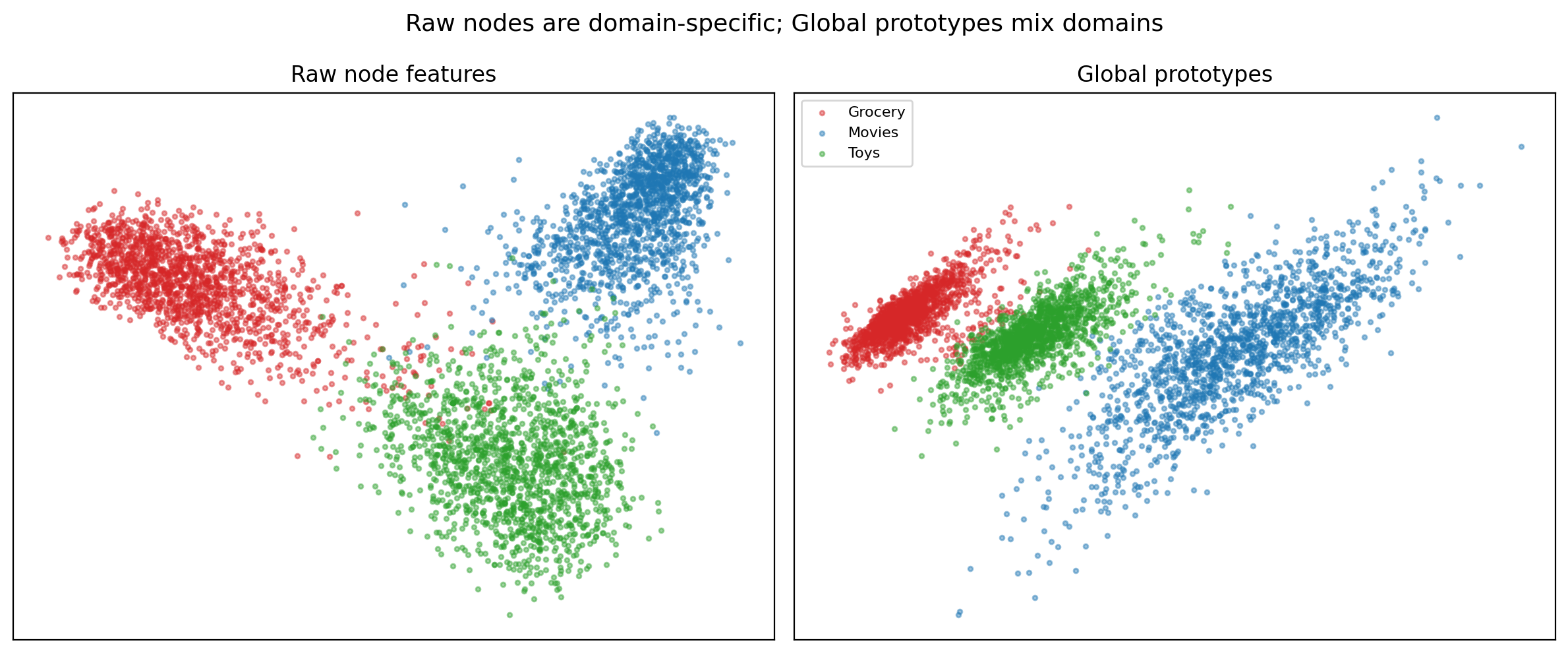}
    \caption{PCA comparison of raw node features and contextualized
    global-token representations across Grocery, Movies, and Toys.}
    \label{fig:app_global_token_pca}
    \Description{Two PCA plots compare domain separation in raw multimodal
    features and contextualized global-token representations.}
\end{figure}

\paragraph{Cross-domain global-token geometry.}
The PCA comparison uses the same evaluated nodes from Grocery, Movies, and
Toys. The raw panel contains cached multimodal node features before hierarchy
retrieval or context propagation. The contextualized panel uses the propagated
output at the retrieved global-anchor position associated with each query.
Because the two representation families have different dimensions and
semantics, PCA~\cite{pearson1901liii} is fitted separately to each family; node indices and domain
colors remain fixed. The visualization is therefore not intended to align
absolute 2D coordinates. Instead, it measures whether domain-specific
separation visible in raw features is reduced after nodes interact with shared
anchors inside the compact context. A more mixed global-token geometry
supports the interpretation that global anchors provide a common semantic
reference space rather than simply reproducing domain-specific raw features.

\section{Complexity Analysis}
\label{app:efficiency}
Let $V=\sum_d|\mathcal{V}_d|$ and $E=\sum_d|\mathcal{E}_d|$ denote the
total numbers of raw nodes and original edges in one three-domain group. Let
$d$ be the frozen feature dimension, $K_1$ and $K_2$ the total numbers of
Layer-1 and Layer-2 anchors, $P$ the number of global anchors, $R$ the
cached PPR candidate count, $K$ the maximum query-context size, $E_c$ the
number of edges in a compact context, $H$ the LLM hidden dimension, and $L$
the number of context-propagation layers. We distinguish the reusable offline
cost from the query-dependent online cost.

\paragraph{Offline hierarchy construction.}
With $I_1$, $I_2$, and $I_g$ Lloyd iterations, the three clustering stages
require
\begin{equation}
O\!\left(I_1VK_1d+I_2K_1K_2d+I_gK_2Pd\right).
\end{equation}
The first term assigns raw nodes to Layer-1 centroids, the second clusters
Layer-1 anchors, and the third constructs the shared global anchors. Generating
synthetic visual features visits the members of each cluster once and costs
$O(Vd)$ across a hierarchy level; encoding the $K_1+K_2+P$ textual summaries
is performed once and cached. Constructing the hierarchical edges requires
$O(V+K_1+mK_2)$ time and storage, because each raw node and Layer-1 anchor has
one parent and each Layer-2 anchor retains at most $m$ global-anchor links.

\paragraph{PPR and bridge preprocessing.}
Let $T_{\mathrm{PPR}}(V,E,R)$ denote the cost of the sparse top-$R$ PPR solver
used to construct all cached candidate lists. The cache occupies $O(VR)$
space. Text and image similarities are evaluated only for the $VR$ retained
pairs, giving $O(VRd)$ time. If two sorted PPR lists of length $R$ are scanned
to compute every overlap score $r_{ij}$, structural filtering has worst-case
cost $O(VR^2)$; since $R=128$ is fixed in our implementation, this remains a
bounded per-node preprocessing cost. Selecting the best $K_b$ bridges with a
size-$K_b$ heap adds $O(VR\log K_b)$ time and $O(VK_b)$ storage. Hence the
complete bridge stage costs
\begin{equation}
O\!\left(T_{\mathrm{PPR}}(V,E,R)+VRd+VR^2+VR\log K_b\right),
\end{equation}
with $O(VR+VK_b)$ cached memory. These operations are performed once per
processed graph and are shared by all epochs, tasks, and target queries.

\paragraph{Online graph-context encoding.}
Because the PPR, hierarchy, and bridge lists are cached, retrieving and
deduplicating a context costs at most $O(K\log K)$. Projecting two modalities
for $K$ items requires $O(KdH)$. When the hidden widths of the reliability and
fusion MLPs are proportional to $H$, gating and fusion cost $O(KH^2)$. Building
and normalizing the compact adjacency costs $O(E_c)$, and one propagation
layer costs $O(E_cH+KH^2)$: the first term is sparse aggregation and the second
is the hidden-state transformation. Center protection, ordering, and layer
normalization add $O(KH)$. The graph-side online complexity per query is thus
\begin{equation}
O\!\left(K\log K+KdH+(L+1)KH^2+LE_cH\right).
\end{equation}
Since $K\leq18$ and $E_c\leq K^2$ in the reported configuration, the online
graph-interface cost is independent of the full graph size after
preprocessing. For link prediction, two endpoint contexts are encoded, which
changes only the constant factor.

\paragraph{LLM computation and memory.}
The frozen backbone still participates in the forward pass and in gradient
propagation to the graph tokens. For a dense Transformer with $L_M$ layers and
a combined textual and graph-token length $T+K$, a standard forward pass has
attention and feed-forward costs of approximately
$O(L_M(T+K)^2H)$ and $O(L_M(T+K)H^2)$, respectively. This term dominates the
bounded graph-side encoder despite the backbone parameters being frozen. The
offline caches require
\begin{equation}
O\!\left(Vd+VR+VK_b+V+K_1+mK_2\right)
\end{equation}
space, excluding raw images and text. During training, the graph encoder stores
$O(LKH+E_c)$ activations per query, while LLM activations dominate peak GPU
memory. Gradient checkpointing reduces this activation memory at the cost of
additional forward recomputation. Overall, \textsc{CHARM} moves graph-size
dependence to a reusable offline stage and keeps the query-time graph encoder
bounded by the fixed context budget.

\end{document}